\documentclass[10pt,twocolumn,letterpaper]{article}

\usepackage[pagenumbers]{cvpr} 
\usepackage[table]{xcolor}
\usepackage{graphicx}
\usepackage{amsmath}
\usepackage{amssymb}
\usepackage{booktabs}
\usepackage{graphics}
\usepackage{multirow}
\usepackage{pifont}
\usepackage{comment}
\usepackage{color}
\usepackage{makecell}
\usepackage{array}

\usepackage[pagebackref,breaklinks,colorlinks]{hyperref}

\usepackage[capitalize]{cleveref}
\crefname{section}{Sec.}{Secs.}
\Crefname{section}{Section}{Sections}
\Crefname{table}{Table}{Tables}
\crefname{table}{Tab.}{Tabs.}

\def\cvprPaperID{2298} 
\def\confName{CVPR}
\def\confYear{2023}

\begin{document}

\title{Rigidity-Aware Detection for 6D Object Pose Estimation}

\author{%
	{Yang Hai $^1$, \quad Rui Song $^1$, \quad Jiaojiao Li $^1$, \quad Mathieu Salzmann $^{2, 3}$, \quad Yinlin Hu $^{4}$} \\
	{\small $^1$ State Key Laboratory of ISN, Xidian University, \quad $^2$ EPFL, \quad $^3$ ClearSpace, \quad $^4$ MagicLeap} \\
}

\maketitle

\begin{abstract}
Most recent 6D object pose estimation methods first use object detection to obtain 2D bounding boxes before actually regressing the pose. However, the general object detection methods they use are ill-suited to handle cluttered scenes, thus producing poor initialization to the subsequent pose network. To address this, we propose a rigidity-aware detection method exploiting the fact that, in 6D pose estimation, the target objects are rigid. This lets us introduce an approach to sampling positive object regions from the entire visible object area during training, instead of naively drawing samples from the bounding box center where the object might be occluded. As such, every visible object part can contribute to the final bounding box prediction, yielding better detection robustness. Key to the success of our approach is a visibility map, which we propose to build using a minimum barrier distance between every pixel in the bounding box and the box boundary. Our results on seven challenging 6D pose estimation datasets evidence that our method outperforms general detection frameworks by a large margin. Furthermore, combined with a pose regression network, we obtain state-of-the-art pose estimation results on the challenging BOP benchmark.

\end{abstract}

\section{Introduction}

\begin{figure}[t]
    \centering
    \setlength\tabcolsep{1pt}
    \begin{tabular}{ccc}
        \includegraphics[width=0.45\linewidth, clip, trim=60 0 20 0]{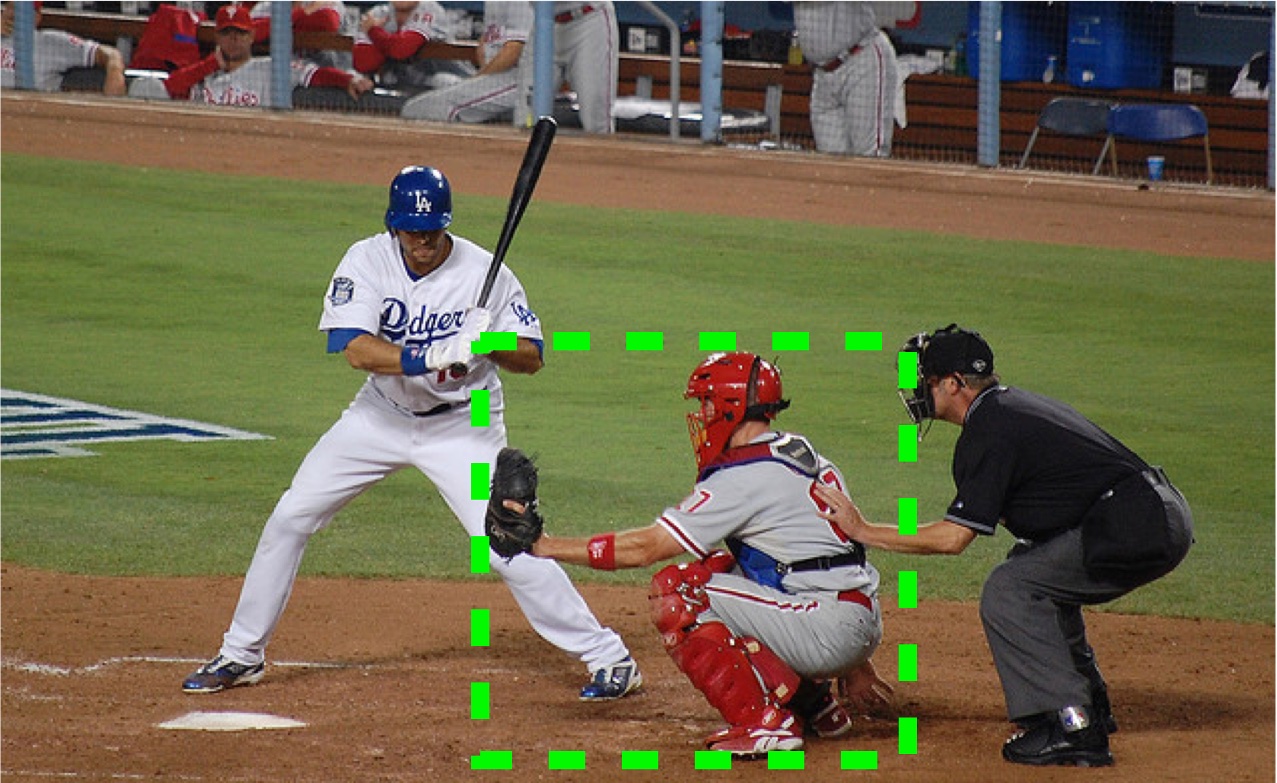} &
        \includegraphics[width=0.45\linewidth, clip, trim=250 260 100 80]{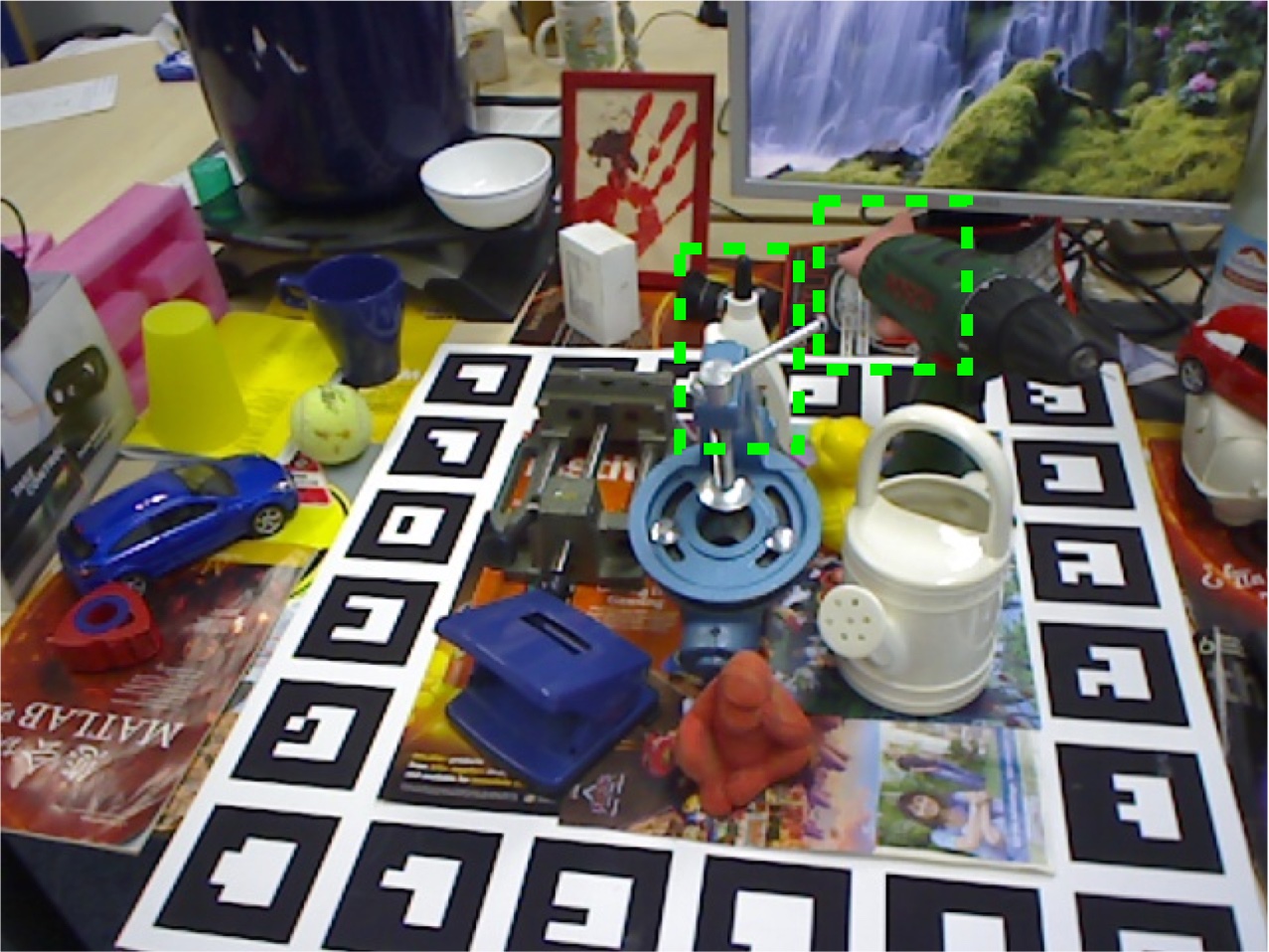} \\
        {\small (a) General detection} & {\small (b) Detection in 6D pose} \\
        \includegraphics[width=0.45\linewidth, clip, trim=250 200 100 80]{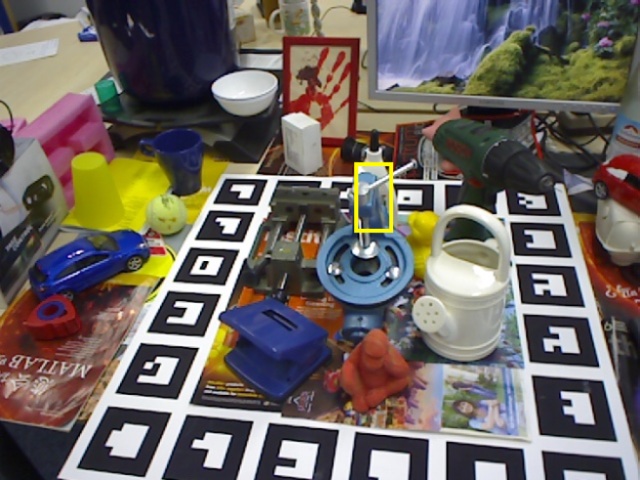} &  
        \includegraphics[width=0.45\linewidth, clip, trim=250 200 100 80]{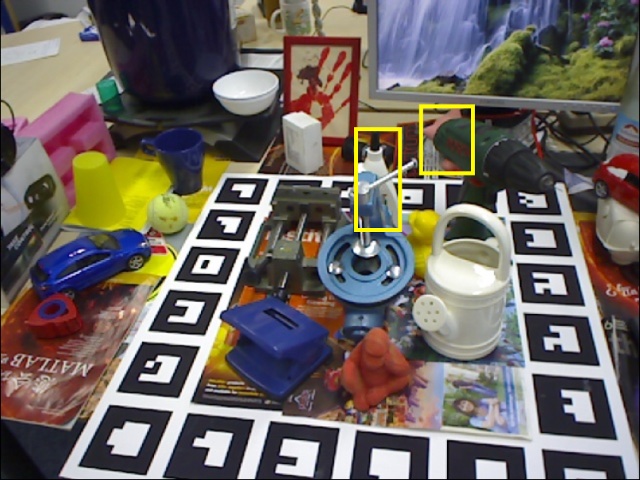} \\
        {\small (c) Baseline detection results} & {\small (d) Our detection results} \\
        \includegraphics[width=0.45\linewidth, clip, trim=250 200 100 80]{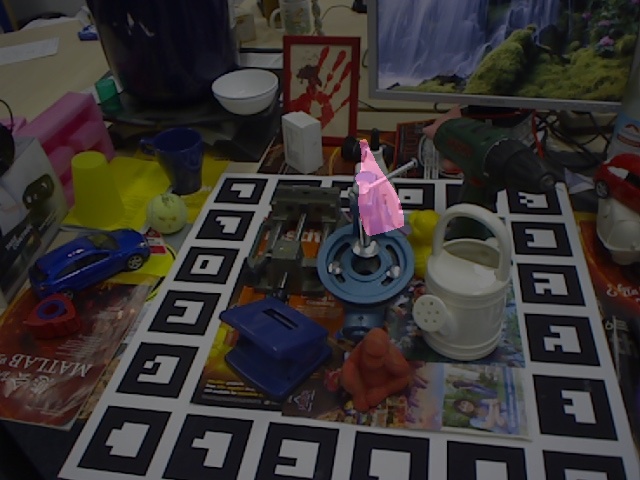} &
        \includegraphics[width=0.45\linewidth, clip, trim=250 200 100 80]{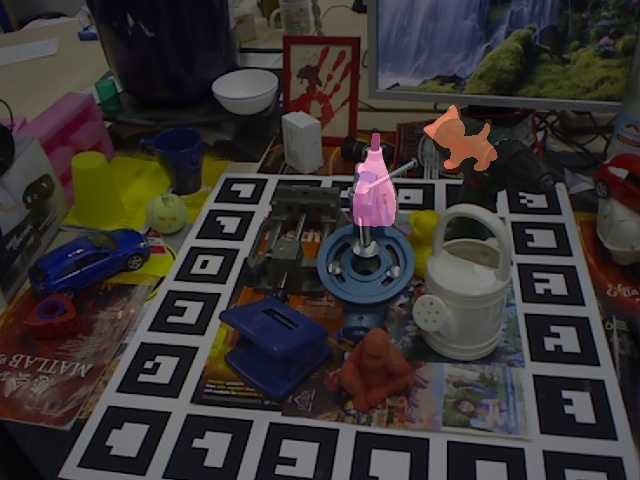} \\
        {\small (e) Baseline pose results} & {\small (f) Our pose results} \\
    \end{tabular}
    \caption{\textbf{The challenges of detection in 6D object pose.} 
    {\bf (a)} The general detection scenario (COCO~\cite{coco}) exhibits small occlusions.
    {\bf (b)} The occlusion problem in 6D object pose, however, is much more severe, {\bf (c)} making the general detection method~\cite{fcosv2} based on center-oriented sampling unreliable (glue) or fail completely (cat). {\bf (d)} By contrast, our new detection strategy is effective in these challenging scenarios, {\bf (e,f)} and provides significantly more robust 2D box initialization for the following 6D regression networks~\cite{pfa}, yielding more accurate pose estimates.
    }
    \vspace{-5pt}
    \label{fig:motivation}
\end{figure}

Estimating the 6D pose of objects, i.e., their 3D rotation and 3D translation with respect to the camera, is a fundamental computer vision problem with many applications in, e.g., robotics, quality control, and augmented reality.
Most recent methods~\cite{zebrapose, deepim, so-pose, surfemb,sc6d, DenseFusion} follow a two-stage pipeline: First, they detect the objects, and then estimate their 6D pose from a resized version of the resulting detected image patches.
While this approach works well in simple scenarios, its performance drops significantly in the presence of cluttered scenes. In particular, and as illustrated in Fig.~\ref{fig:motivation}, we observed this to be mainly caused by detection failures. 

\begin{figure}
    \centering
    \setlength\tabcolsep{2pt}
    
\begin{tabular}{ccccccl}
  \includegraphics[width=0.15\textwidth]{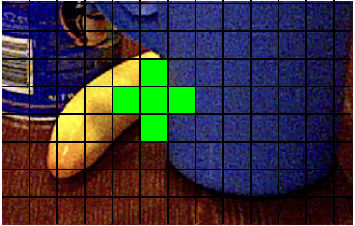} &
  \includegraphics[width=0.15\textwidth]{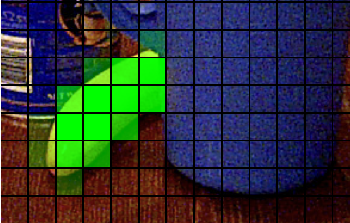} &
  \includegraphics[width=0.15\textwidth]{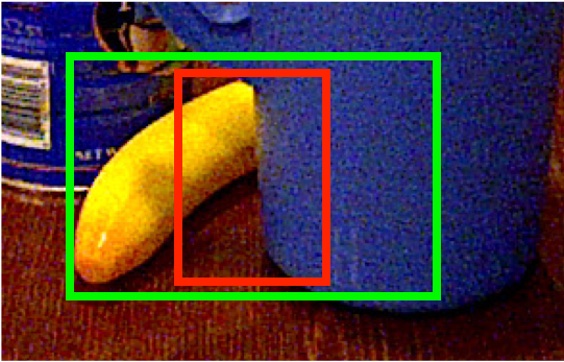} \\
  \includegraphics[width=0.15\textwidth]{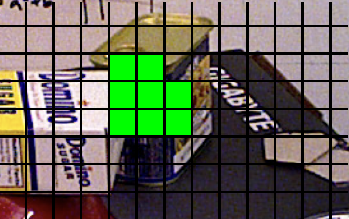} &
  \includegraphics[width=0.15\textwidth]{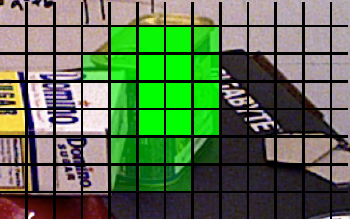} &
  \includegraphics[width=0.15\textwidth]{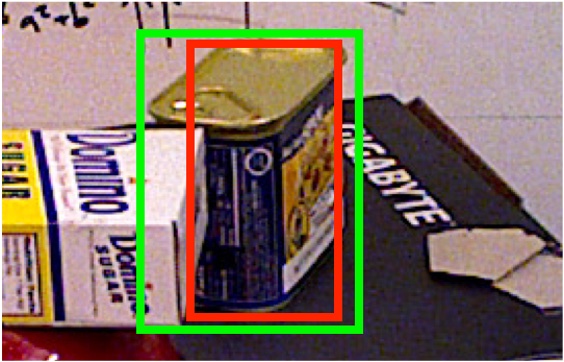} \\
  {\small (a) Baseline strategy} & {\small (b) Our strategy} & {\small (c) Detection results}\\
\end{tabular} 
    \vspace{-3pt}
    \caption{\textbf{Detecting rigid objects in cluttered scenes.}
    {\bf (a)} The standard strategy~\cite{ATSS} chooses positive samples (green cells) around the object center, thus suffering from occlusions. {\bf (b)} Instead, we propose to use a visibility-guided sampling strategy to discard the occluded regions and encourage the network to be supervised by all visible parts. The sampling probability is depicted by different shades of green. {\bf (c)} Our method (green boxes) yields more accurate detections than the standard strategy (red boxes).
    }
    \label{fig:atss_vs_ours_demo}
\end{figure}

Specifically, most 6D pose estimation methods rely on standard object detection methods~\cite{ATSS, fcosv1, fcosv2, PAA, faster-rcnn, maskrcnn}, which were designed to handle significantly different scenes than those observed in 6D object pose estimation benchmarks, typically with much smaller occlusions, as shown in Fig.~\ref{fig:motivation}(a). 
Because of these smaller occlusions, standard detection methods make the assumption that the regions in the center of the ground-truth bounding boxes depict the object of interest, and thus focus on learning to predict the bounding box parameters from samples drawn from these regions only.
However, as shown in Fig.~\ref{fig:atss_vs_ours_demo}, this is ill-suited to 6D pose estimation in cluttered scenes, where the center of the objects is often occluded by other objects or scene elements. 

To handle this, we propose a detection approach that leverages the property that the target objects in 6D pose estimation are rigid. For such objects, any visible parts can provide a reliable estimate of the complete bounding box. We therefore argue that, in contrast with the center-based sampling used by standard object detectors, any, and only feature vectors extracted from the visible parts should be potential candidates of positive samples during training.


In principle, modeling the visibility could be achieved by annotating segmentation masks for all objects. This process, however, is cumbersome, particularly in the presence of occlusions by scene elements, and would limit the scalability of the approach. 
Instead, we therefore propose to compute a probability of visibility based on a minimum barrier distance between any pixel in a bounding box and the box boundary.
We then use this probability to guide the sampling of candidates during training, thus discarding the occluded regions and encouraging the network to be supervised by all visible parts. 
Furthermore, to leverage the reliability of local predictions from most visible parts during inference, we collect all candidate local predictions above a confidence threshold, and combine them by a simple weighted average, yielding more robust detections.

We demonstrate the effectiveness of our method on seven challenging 6D object pose estimation datasets, on which we consistently and significantly outperform all detection baselines. Furthermore, combined with a 6D pose regression network, our approach yields state-of-the-art object pose results. 

\section{Related Work}

\noindent\textbf{Object pose estimation}, whose goal is to estimate the 3D rotation and 3D translation of a target object with respect to the camera, nowadays typically involves a pose regression network to establish 3D-to-2D correspondences~\cite{segdriven, epos, pvnet,single_stage_hu, wdr, pfa, ncf}. These correspondences then act as input to a perspective-n-points solver (PnP)~\cite{epnp} to compute the final 6D object pose. The current state-of-the-art methods~\cite{pix2pose, cdpn, surfemb, sc6d, zebrapose, cosypose, dpodv2,so-pose, gdr_net} virtually all use a 2D object detector to allow the following pose regression networks to focus on a region of interest (RoI), thus yielding more accurate poses. 

While this is effective when detection is successful, the pose accuracy deteriorates significantly in case of missing or inaccurate detections. In particular, 6D pose estimation frameworks typically use standard object detectors that, as shown in Figs.~\ref{fig:motivation} and~\ref{fig:atss_vs_ours_demo}, often fail in cluttered scenes such as those of standard 6D pose estimation benchmarks as they were not designed to handle such situations. To handle this, we propose a rigidity-aware detection method that leverages the target properties. As shown by our results, it yields significant better RoIs for 6D object pose estimation.

\noindent\textbf{Object detection}, whose goal is to extract accurate 2D bounding boxes for all objects in a scene, has been widely studied in 2D computer vision. Existing methods follow one of two main strategies: two-stage or one-stage detection. Two-stage detectors first employ a region proposal network~\cite{faster-rcnn, maskrcnn} to generate bounding box candidates, which are then processed by a classification and refinement network to remove false positives and adjust the bounding boxes position and size~\cite{faster-rcnn,cascade-rcnn,maskrcnn}. 
Although this strategy is accurate in general, it is costly and inefficient in practice.

One-stage detectors tackle this by replacing the region proposal network with a pre-defined set of anchors at every spatial location in the encoder's final feature map~\cite{retinanet,fcosv1,yolov1}. Unfortunately, this suffers from the presence of many negative samples among the anchors. While this can be addressed to some degree by FocalLoss~\cite{retinanet,fpn}, early single-stage detectors did not reach the accuracy of two-stage ones.

This was addressed in~\cite{ATSS} 
via a simple yet effective strategy to sample positive candidates in a one-stage detector.
Most recent detection methods follow similar strategies~\cite{fcosv2, PAA, autoassign, OTA, TTF, yolov2, yolov3}, and now achieve better accuracy than two-stage methods while being more efficient. 

Nevertheless, while these methods work well on standard object detection benchmarks, they suffer from the heavy occlusions present in 6D pose estimation ones. Here, we therefore propose a new strategy dedicated to detecting rigid objects, and show that it outperforms standard detectors by a large margin in the context of 6D pose estimation.




\begin{figure}[t]
    \centering
    \includegraphics[width=1.0\linewidth]{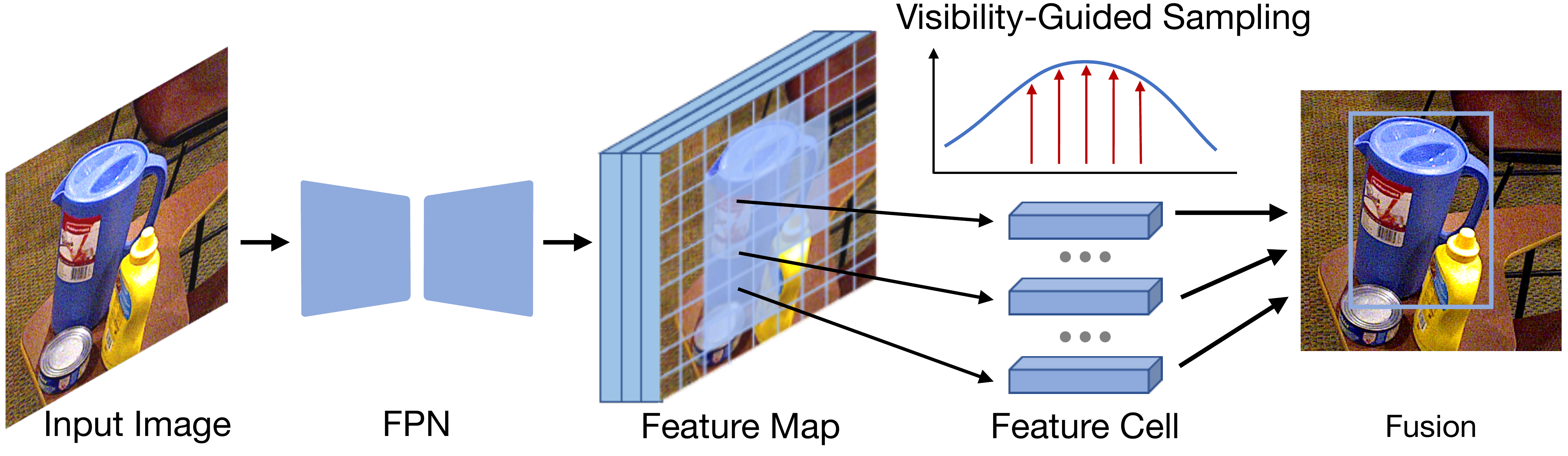}
    \caption{{\bf Overview of our detection approach.}
    We use a general Feature Pyramid Network (FPN) as our backbone. We first compute a probability of visibility for every local area within the bounding box, which we use to guide the sampling of positive cells during training, without any mask annotations. Finally, during inference, we combine all the local candidate predictions to obtain a more robust final result.
    }
    \label{fig:training}
\end{figure}


\section{Approach}

Given an RGB image depicting rigid objects, our goal is to estimate the 2D bounding box of each potential target for the subsequent pose regression network. 
To address this, we propose to leverage the fact that, in the context of 6D object pose estimation, we observe rigid targets.
In this section, we first briefly review the problem of positive sampling in object detection and analyze the influence of the objects' rigidity in 6D object pose scenarios. We then explain how we compute object foreground probabilities without having access to ground-truth masks, and introduce a positive sampling strategy based on these probabilities. Finally, we propose a box fusion strategy to improve detection robustness. Fig.~\ref{fig:training} provides an overview of our detection approach.

\subsection{Analysis of Rigidity in Detection}
\label{sec:analysis}

Modern single-stage object detectors~\cite{fcosv1, ATSS, PAA, autoassign,retinanet} rely on a Feature Pyramid Network (FPN)~\cite{fpn} that outputs scale-rich feature maps. Each feature vector is taken as a training sample and further processed by a classification branch and a regression branch. Training the detector thus first requires defining positive and negative samples for each annotated object instance. The positive samples are then encouraged to be classified as the instance's category, whereas the negative samples should be predicted as background. Furthermore, the positive samples should regress the instance's bounding box parameters. Since during training a single instance is associated with multiple positive samples, at inference multiple samples will be activated for a potential target. Most methods then use the standard Non-Maximum Suppression (NMS) as a post-processing stage to obtain the final result.

Key to the success of this general framework is the selection
of positive samples during training.
The standard approach to sampling positive features during training consists of assuming that the regions in the center of the ground-truth bounding boxes depict the object.
However, in the context of 6D pose estimation, this center assumption is often violated because of the large occlusions that occur in cluttered scenes. More importantly, it does not account for the fact that, for rigid target objects, all visible object parts can provide a reliable prediction of the entire bounding box.

\begin{figure}
\setlength\tabcolsep{1pt}
  \begin{tabular}{cc}
    \includegraphics[width=0.49\linewidth]{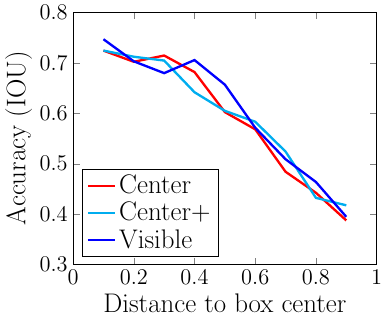} &
        \includegraphics[width=0.49\linewidth]{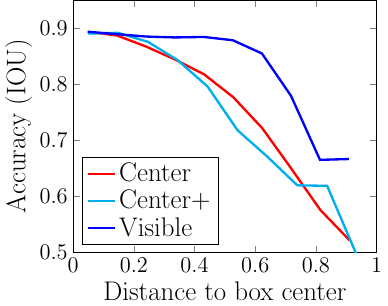} \\
    {\small (a) General (COCO)}        & {\small (b) Rigid (YCB)}
  \end{tabular}
  \vspace{-3pt}
  \caption{{\bf Analysis of rigidity in detection.}
    We show the testing accuracy of different sampling strategies w.r.t. different local predictions during training on the typical general object dataset (COCO~\cite{coco}) and on the typical 6D object pose dataset (YCB~\cite{posecnn}). We report the results of FCOSv2~\cite{fcosv2} (Center), ATSS~\cite{ATSS} (Center+), and a strategy exploiting all the  candidates in the ground-truth mask (Visible). The horizontal axis represents the normalized distance of a local prediction to the box center. Although the accuracy of different strategies is similar on COCO, the visibility-guided sampling is much more accurate on YCB, even when the local predictions come from non-center areas, thanks to the rigidity of the target objects.
    }
  \label{fig:rigid_discussion}
\end{figure}

To evidence this, we train the same FPN network with different sampling strategies on the general COCO dataset~\cite{coco} and on the typical 6D object pose YCB~\cite{posecnn} dataset, respectively. We first evaluate two baseline strategies, consisting of sampling a fixed number of positive cells from the center region in the ground-truth bounding box (FCOSv2~\cite{fcosv2}), and of an adaptive center-based sampling strategy across all pyramid feature levels (ATSS~\cite{ATSS}). Furthermore, we evaluate a sampling strategy that randomly chooses 10 positive cells within the ground-truth object mask.

Fig.~\ref{fig:rigid_discussion} depicts the average test accuracy of different local predictions obtained with these sampling strategies as a function of the distance of the prediction to the true bounding box center. On the general COCO dataset, the accuracy deteriorates as the distance increases regardless of which sampling strategies was used during training. This comes from the diversity of the object types in COCO, which includes many non-rigid objects and a wide variety of instances with the same object type, making the object center a more reliable predictor of the bounding box.
On YCB, the accuracy of the centered-based strategies also deteriorates quickly as the distance increases, since most non-center area were not involved during training. However, thanks to the rigidity of the YCB targets, the visibility-guided strategy yields more accurate local predictions,
even for those that are farther away from the center area.

\begin{figure}[t]
    \centering
    \setlength\tabcolsep{1pt}
    \begin{tabular}{ccc}
    \includegraphics[width=0.32\linewidth]{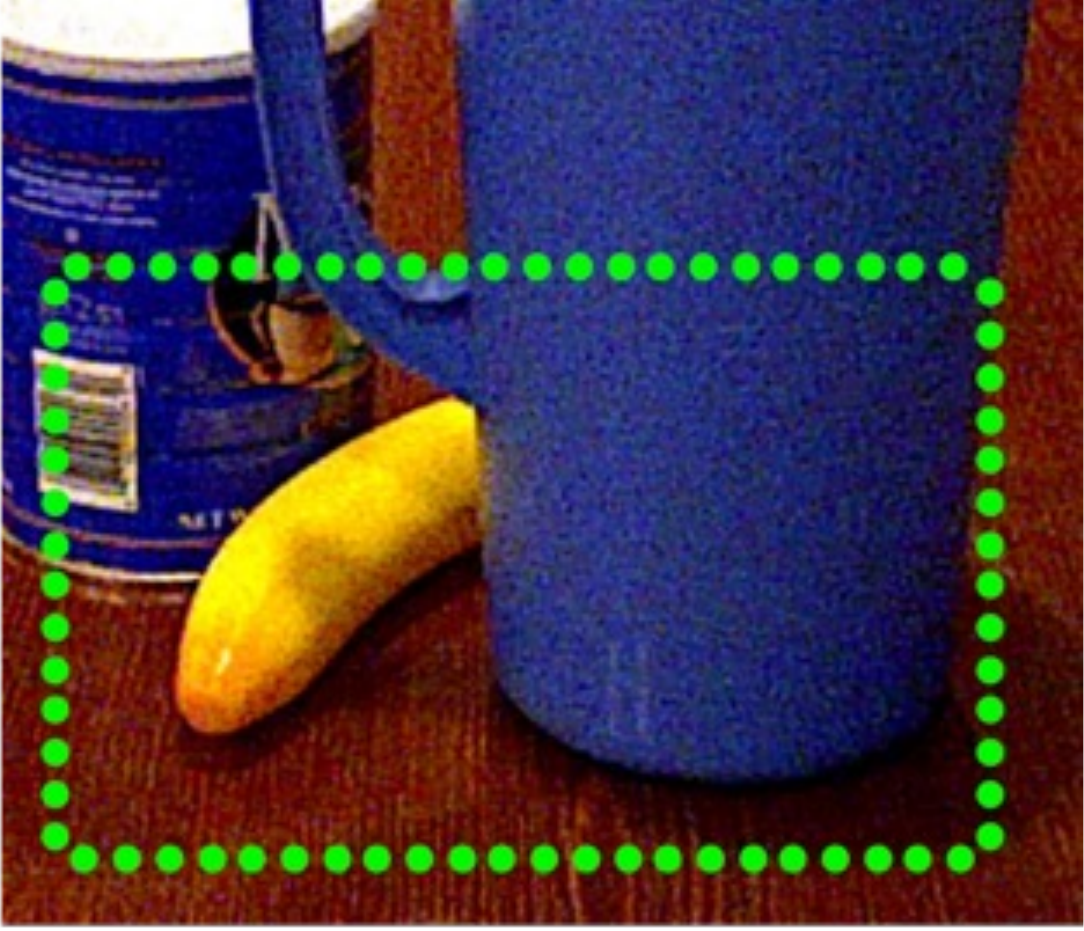} &
    \includegraphics[width=0.32\linewidth]{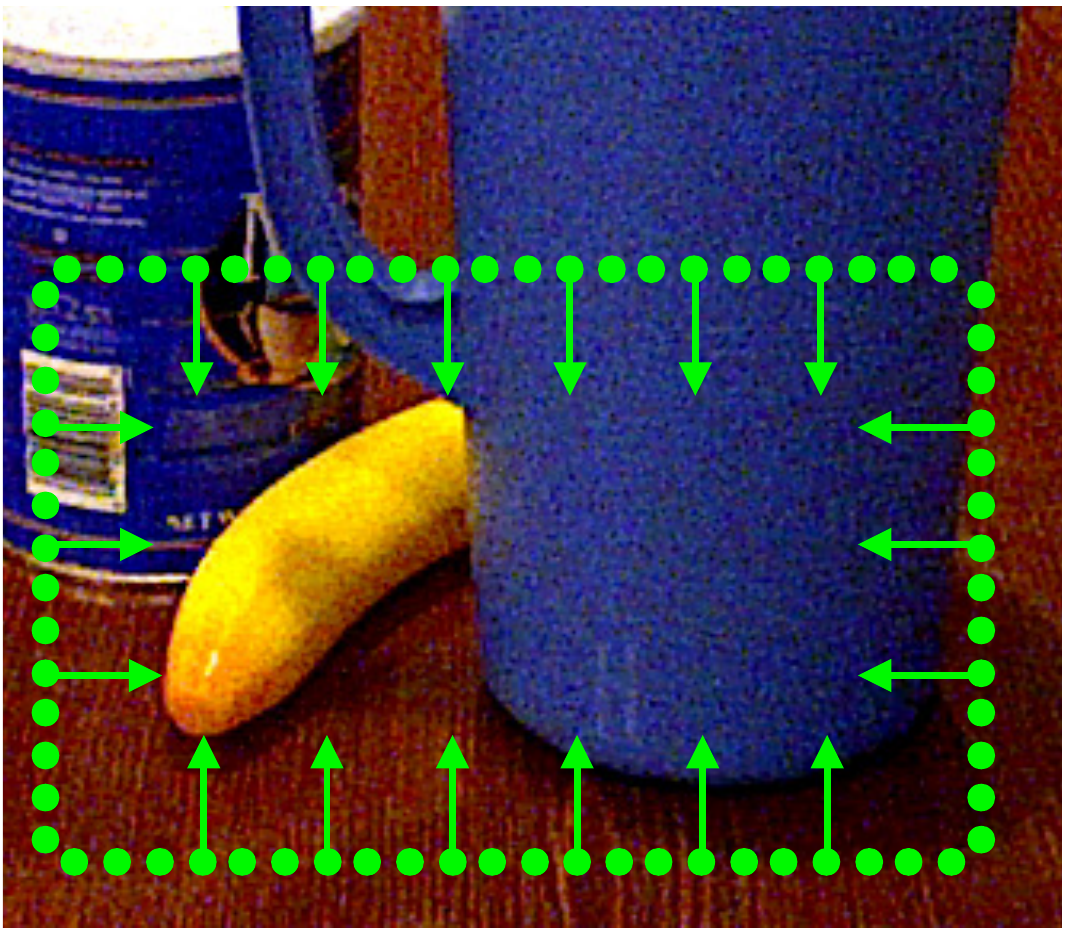} &
    \includegraphics[width=0.32\linewidth]{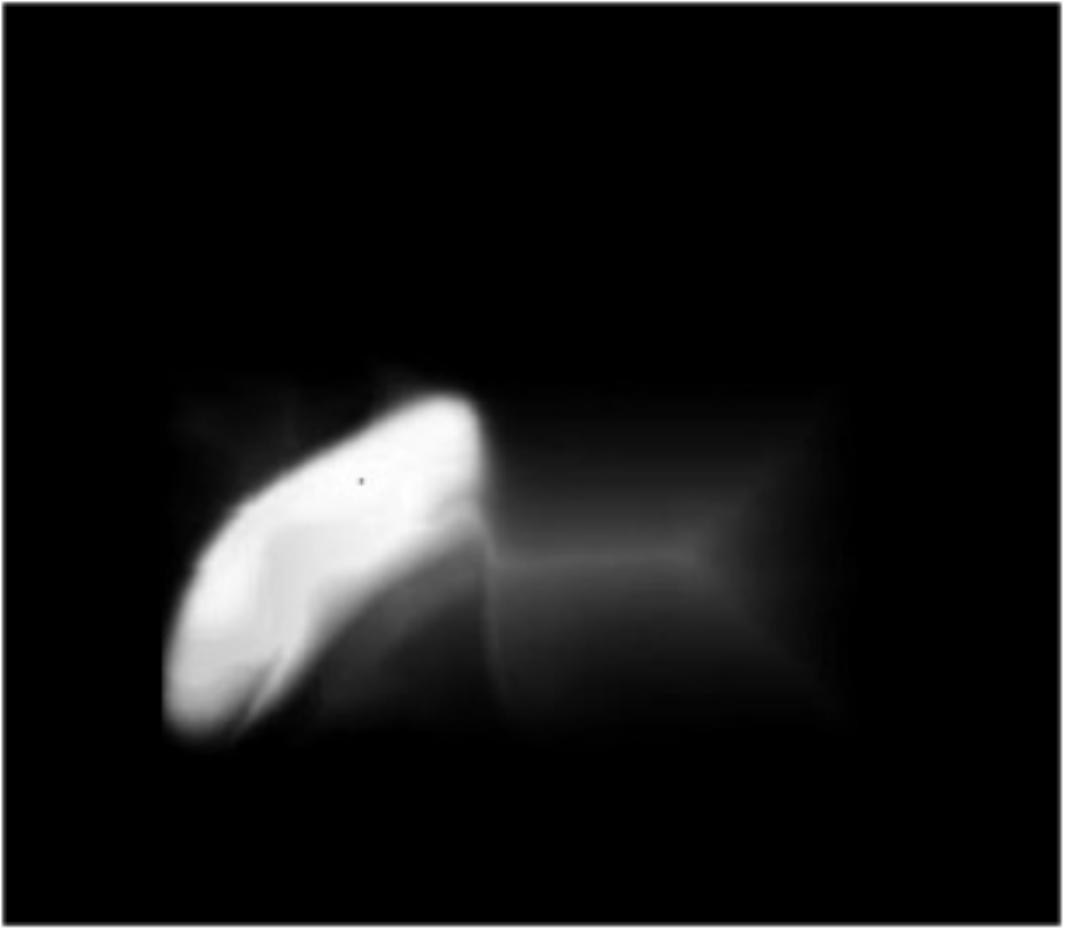} \\
    {\small (a) Initial seeds} & {\small (b) Seeds growing} & {\small (c) Distance map} \\
\end{tabular}

    \caption{\textbf{Visibility modeling without mask annotation.}
    {\bf (a)} We first place a set of seeds on the bounding box boundaries, {\bf (b)} then compute a minimum barrier distance between every pixel within the box and the seeds,
    {\bf (c)} obtaining a distance map 
    from which we build a probability of visibility for every local object part.
    }
    \label{fig:distance_transform}
\end{figure}

\subsection{Visibility-Guided Sampling}
\label{sec:guided_sampling}

The strategy used in the previous experiment relies on the ground-truth object mask during training. However, such masks are typically not available and expensive to obtain, particularly in the presence of occlusions with scene elements. To avoid requiring such masks, we compute an approximate measure of visibility for each pixel in the ground-truth object bounding box.

To this end, let $\mathcal{I} \in \mathbb{R}^{H\times W \times 3}$ be an image patch obtained by cropping a ground-truth object bounding box.
We then create a seed set $\mathcal{S}=\{s_1,\cdots,s_m\}$ of 2D positions in the image patch by uniformly sampling the patch boundary with a fixed step size~\cite{zhang2015minimum, wei2012geodesic}. Our method then builds on the intuition that these seeds will typically \emph{not} belong to the target object. Therefore, the visible object pixels should significantly differ from the seeds, which we encode using an online distance transform.

Specifically, we compute the distance from each pixel within the patch to its nearest seed. For a pixel $p$ and with a generic distance metric, i.e., without assuming the use of the Euclidean distance, this can be expressed as
\begin{equation}
  {\cal D}(p) = \min_{s \in \mathcal{S}} \mathcal{D}(p, s)\;,
\end{equation}
where $\mathcal{D}(p, s)$ encodes the distance between pixel $p$ and seed $s$. Such a distance can in general be expressed as
\begin{equation}
  {\cal D}(p, s) = \min_{\tau \in \prod_{\{p, s\}}}\mathcal{H}(\tau ),
  \label{eqn:distance_transform}
\end{equation}
where $\tau$ is a path connecting pixel $p$ and seed $s$, $\mathcal{H}(\tau )$ is the cost of path $\tau$, and $\prod_{\{p, s\}}$ is the set containing all possible paths connecting $p$ and $s$. 

Here, we define the cost $\mathcal{H}(\tau )$ as the minimum barrier distance~\cite{minimum_distance,minbarrier}, i.e.,
\begin{equation}
  \mathcal{H}(\tau )= \mathcal{B}(\mathcal{I}, \tau) + \alpha \cdot d(\tau_0, \tau_1),
  \label{eqn:mbd}
\end{equation}
where $\tau_0$ and $\tau_1$ are the path's starting and ending point, respectively, $d(\tau_0, \tau_1)$ is the Euclidean distance between these two points, and
\begin{equation}
    \mathcal{B}(\mathcal{I}, \tau) = \mathop{max}_{i=1}^{3}(\mathop{max}\limits_{t=0}^{1}\mathcal{I}_i(\tau_t) - \mathop{min}\limits_{t=0}^{1}\mathcal{I}_i(\tau_t)),
\end{equation}
with $\mathcal{I}_i(\tau_t)$ the intensity of the $i^{th}$ channel at a pixel $\tau_t$ along the path. We set the balance factor $\alpha=0.1$ in our experiments, which makes the distance rely mainly on the difference between the maximum and minimum pixel value along the path, thus improving the robustness to different illumination conditions~\cite{minimum_distance}.

The resulting distance can be computed efficiently using a fast minimum-barrier-distance solver~\cite{minbarrier}, which lets us generate the corresponding distance transform map ${\cal D}(p)$. Fig.~\ref{fig:distance_transform} illustrates the procedure discussed above, showing that it correctly reflects the object visibility.

In essence, our distance maps provide us with soft visibility masks for the target objects. We then use these soft masks to sample positive cells in a single-stage detection framework, as discussed in Section~\ref{sec:analysis}.

To this end, for every cell $c$ in every feature map extracted by the FPN module, we compute a visibility score 
that $c$ belongs to the object as
\begin{equation}
  \centering
  {\cal V}(c) = \frac{\bar{\cal D}(c)}{\max_{f\in \mathcal{F}} \bar{\cal D}(f)},
  \label{eqn:distance_to_pro}
\end{equation}
where $\bar{\cal D}(c)$ averages the distance map values of all the pixels encompassed by cell $c$, and $\mathcal{F}$ is the set of all cells in the feature map of interest. We then only consider the cells such that ${\cal V}(c) > \mathcal{T}$ as candidate positives, and use $\mathcal{T}=0.25$ in our experiments.

Note, however, that using all the cells with ${\cal V}(c) > \mathcal{T}$ as positives would result in training being dominated by larger objects. To prevent this, we randomly select $k=10$ cells for each object instance according to ${\cal V}(c)$.
For the instances containing less than $k$ foreground cells, we randomly sample existing ones multiple times to nonetheless obtain $k$ positive samples. We then discard the cells not chosen as positive samples yet still having a visibility score larger than the threshold $\mathcal{T}$ from the classification and box regression process, to avoid providing the network with potentially inconsistent supervision signal.



\begin{figure}[t]
    \centering
    \setlength\tabcolsep{2pt}
    
\setlength\tabcolsep{1pt}
\begin{tabular}{ccccc}
  \includegraphics[width=0.24\linewidth, clip, trim=100 0 700 400]{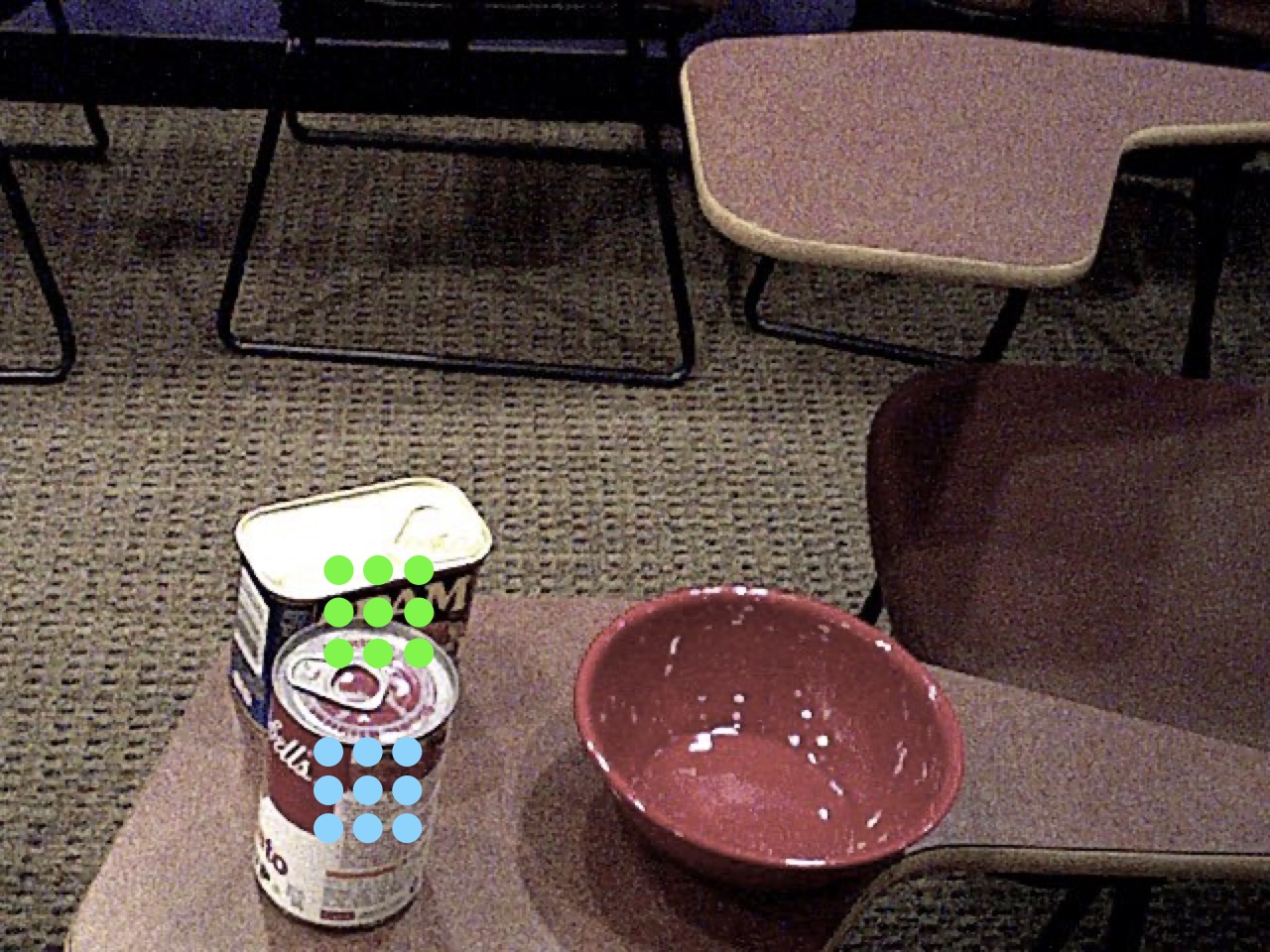} &
  \includegraphics[width=0.24\linewidth, clip, trim=100 0 700 400]{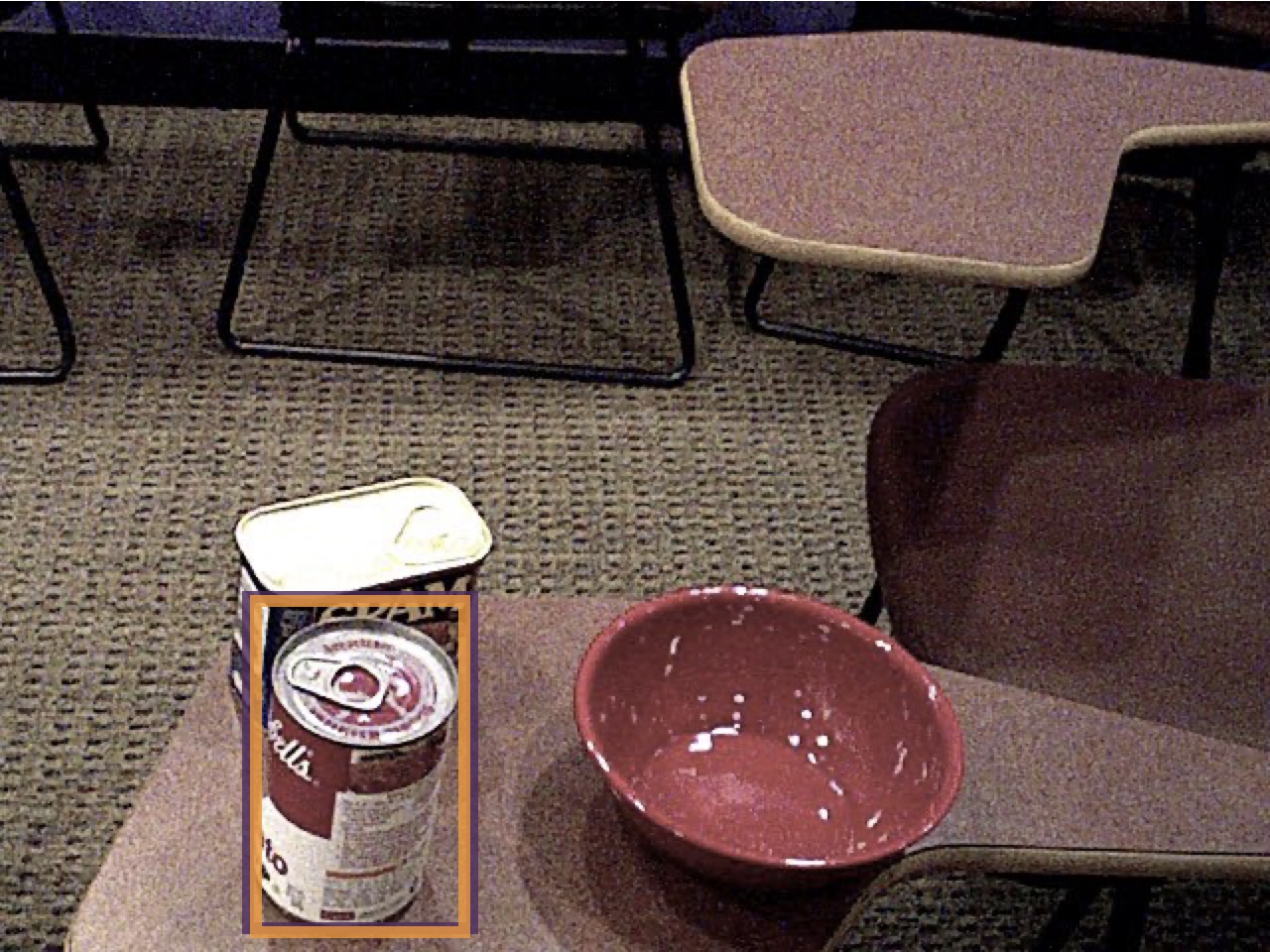}&
  \includegraphics[width=0.24\linewidth, clip, trim=100 0 700 400]{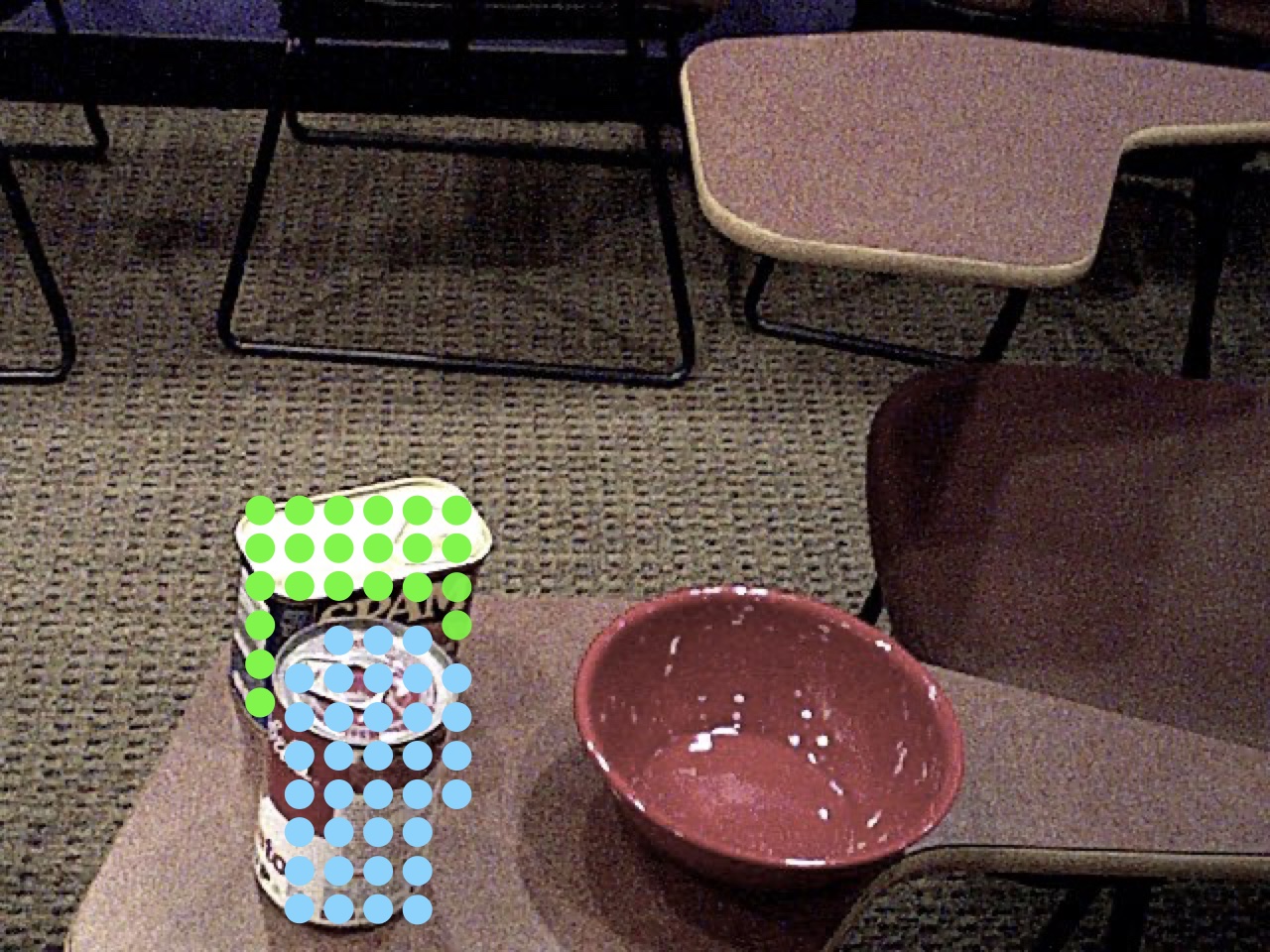} &
  \includegraphics[width=0.24\linewidth, clip, trim=100 0 700 400]{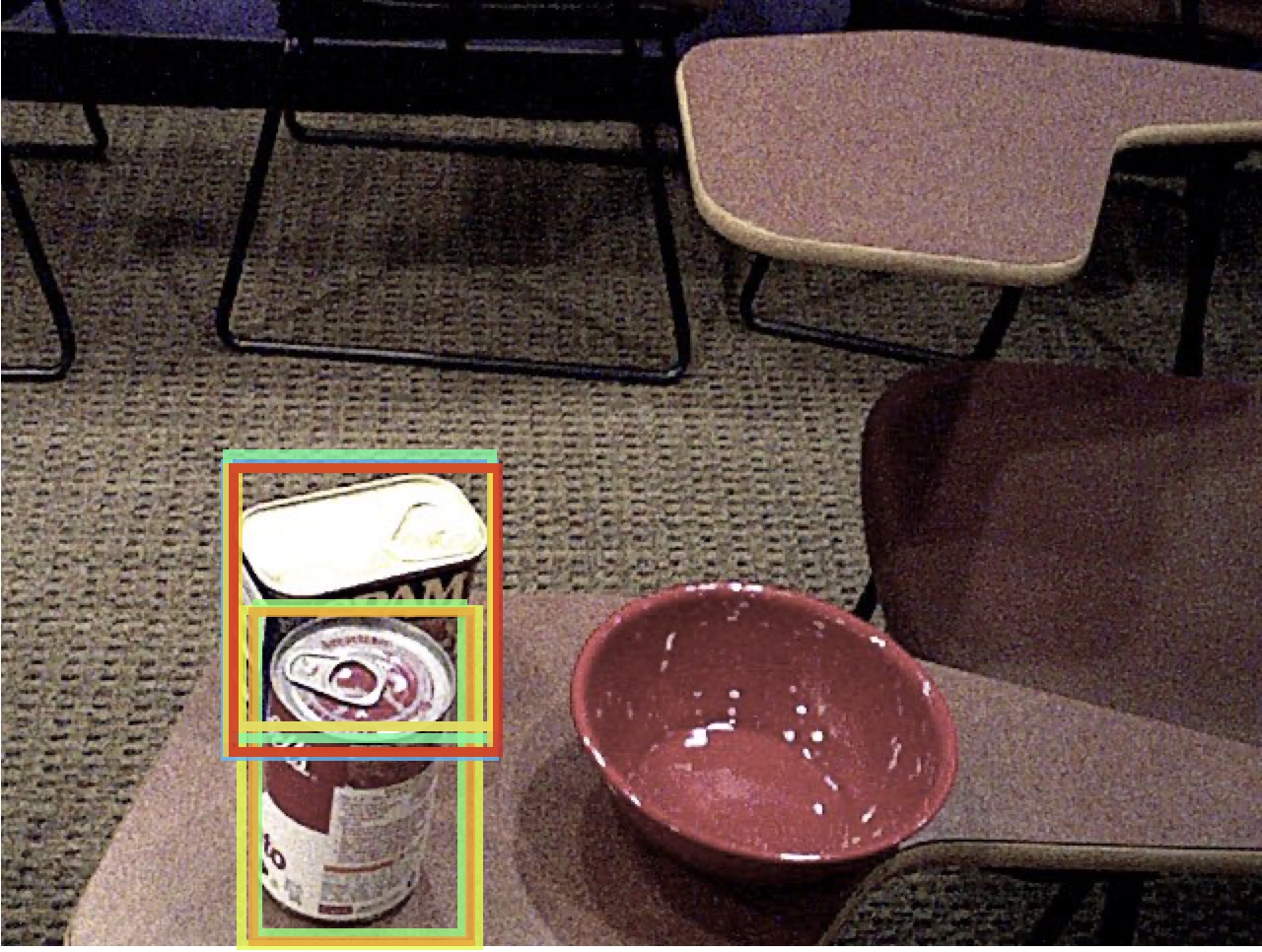} &
  \includegraphics[width=0.017\linewidth]{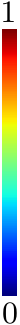} \\
  \multicolumn{2}{c}{\small (a) The standard strategy} & \multicolumn{2}{c}{\small (b) The proposed strategy} & ~\\
\end{tabular}

    \caption{{\bf Robustness of different strategies.}
    The left and right parts of {\bf (a)} and {\bf (b)} show the sampling strategy during training and all the valid local predictions before fusion during inference, respectively. The standard center-based sampling strategy suffers from occlusions, as evidenced by the lack of valid predictions for the upper box. Additionally, it generates candidate predictions with large differences in confidence values (as shown by the color difference for the lower box). By contrast, our strategy is robust to occlusions and yields more candidate predictions with high confidence, which can be combined to obtain better results.
  }
  \label{fig:fusion_compare}
\end{figure}

\begin{table*}[t]
    \centering

\begin{tabular}{l|ccccccc|c}
    \toprule
    Method                      & LM-O          & T-LESS        & TUD-L         & IC-BIN        & ITODD         & HB            & YCB         & Avg.\\
    \midrule
    \textbf{Ours}              & \textbf{67.5} & \textbf{79.8}& \textbf{86.6}& \textbf{63.8}& \textbf{48.6}& \textbf{73.5}& \textbf{85.0}& \textbf{72.1}   \\
    FCOSv2~\cite{fcosv2}          & 57.0         & 75.0         & 86.0         & 27.2         & 30.4         & 60.4         & 80.0         & 66.7     \\
    Mask R-CNN~\cite{maskrcnn}  & 56.6         & 69.3         & 82.6         & 40.1         & 36.5         & 63.5         & 74.5         & 60.5     \\ 
    \bottomrule
    
\end{tabular}
    \caption{{\bf Detection comparison on different 6D object datasets.}
    Our method achieves much better accuracy than the baseline methods on these BOP datasets, demonstrating the effectiveness of our approach at detecting rigid objects in cluttered 6D pose estimation scenarios.
    }
    \label{tab:detect_compare1}
\end{table*}

\subsection{Fusion During Inference}
\label{sec:fusion}

As discussed in Section~\ref{sec:analysis}, during inference, each object instance typically receives multiple box predictions.
On the general COCO dataset, Non-Maximum Suppression~\cite{fcosv1, fcosv2, ATSS} is typically the method of choice to select a single box, choosing the candidate with the maximum confidence within a local area. This strategy builds on the assumption that only a small region within the box, typically near the box center, can provide a prediction with high precision, as shown in Fig.~\ref{fig:rigid_discussion}(a). In the 6D pose estimation setting, however, all visible parts can provide almost equally-accurate predictions, thanks to the rigidity of the targets, as shown in Fig.~\ref{fig:rigid_discussion}(b). 

We therefore propose to combine all the candidate boxes in a neighborhood to obtain a more accurate result. To this end, we let the feature cells predict an additional confidence value, representing how precise the predicted box is. We then cluster the different local predictions that have the same local maximum and assign them to the same object instance. This strategy is similar to the NMS one, but without any candidate suppression. We then compute a simple weighted sum to combine all the candidate local predictions within the same cluster, with weights based on the predicted confidence values.
Fig.~\ref{fig:fusion_compare} demonstrates the advantages of this strategy.

\subsection{Implementation Details}

As mentioned above, we use the same FPN architecture as most state-of-the-art single-stage frameworks~\cite{ATSS, fcosv2,  PAA, autoassign}. We define the confidence value as the IOU between the predicted box and the ground-truth one.
We then train our model with a combined loss function
\begin{equation}
  \centering
  \mathcal{L} = \mathcal{L}_{cls}(\theta, g) + \mathcal{L}_{reg}(\theta, g)
  + \mathcal{L}_{iou}(\theta, g),
  \label{eqn:loss_func}
\end{equation}
where $\theta$ denotes the model parameters and $g$ encodes the ground-truth boxes.
$\mathcal{L}_{cls}$ is the focal loss for classification, $\mathcal{L}_{reg}$ is the box regression loss, and $\mathcal{L}_{iou}$ is the
binary cross entropy between the predicted IOU and the ground-truth IOU for confidence prediction.
We use GIOU~\cite{giou} loss for $\mathcal{L}_{reg}$ in our implementation.

During training, we first assign every instance to one pyramid level on FPN according to the object size, similarly to~\cite{fcosv1}. We then compute our distance map on the fly within the annotated bounding box and use it to guide the positive sampling as discussed above.
During inference, we use a threshold of 0.05 based on the classification score to remove most of  the noise from the background before the clustering and fusing the boxes as discussed in Section~\ref{sec:fusion}.

\section{Experiments}
\label{sec:experiment}

In this section, we systematically study our detection method in 6D object pose estimation scenarios. We first compare its detection performance with other detection baselines in Section~\ref{sec:exp_detection}, and then examine its effect when used as bounding box initialization for different pose regression networks in Section~\ref{sec:exp_pose}. Our source code is available at \url{https://github.com/YangHai-1218/RADet}.

\noindent \textbf{Experimental settings.}
We evaluate our method on seven core datasets from the BOP benchmarks~\cite{bop}, including LM-O~\cite{lmo}, T-LESS~\cite{t-less}, TUD-L~\cite{bop}, IC-BIN~\cite{icbin}, ITODD~\cite{itodd}, HB~\cite{hb}, and YCB~\cite{posecnn}, which are standard benchmarks for 6D object pose estimation. Most of the datasets have both real images and synthetic ones generated by physically based rendering (PBR)~\cite{blenderproc} for training, and another split of real images for testing. We use mixed data for training by default. However, for LM-O, IC-BIN, ITODD, and HB, we have only 50k synthetic images for training. As such, we train models only on synthetic images on these datasets.

For a fair comparison with other detection methods, we use the same training setting for both our method and all the competitors unless otherwise stated. We use a ResNet-50 backbone~\cite{resnet} with pre-trained weights from ImageNet~\cite{imagenet}, a batch size of 16, and an input image resolution fixed at 640$\times$480. We train all the models
with the SGD optimizer for 90k iterations, using an initial learning rate of 0.01 with a decay ratio of 0.1 after 60k and 80k iterations, respectively.


\noindent \textbf{Evaluation metrics.}
We report numbers in the standard metric AP for detection results~\cite{fcosv1,ATSS,retinanet}, which is the average value of different AP values obtained with an IOU threshold between the ground truth box and the predicted one ranging from 0.5 to 0.95. For a detailed study, we also report AP$_{50}$ and AP$_{75}$, which use an IOU threshold of 0.5 and 0.75, respectively.

For 6D pose estimation, we report the three standard metrics used in the BOP benchmarks, including the Visible Surface Discrepancy (VSD), the Maximum Symmetry-aware Surface Distance (MSSD), and the Maximum Symmetry-aware Projection Distance (MSPD)~\cite{bop}. In essence, these metrics differ in the strategies they use to measure the distance between the ground-truth pose and the estimated one. We refer the readers to~\cite{bop} for their detailed definitions. We report the average numbers of these three metrics in some of our evaluations to save space, and encourage the reader to check the appendix for the detailed numbers of each metric.

\begin{table}[t]
  \setlength{\fboxrule}{0pt}
  \begin{center}

\begin{tabular}{lccc}
    \toprule
    Method                              & AP             & AP$_{50}$     & AP$_{75}$     \\
    \midrule
    \textbf{Ours}                       & {\bf 85.0}    & {\bf 99.4}    & {\bf 97.4}        \\
    PAA~\cite{PAA}                      & 83.5            & 98.3         & 93.2         \\
    AutoAssign~\cite{autoassign}        & 83.3            & 98.1         & 91.7         \\
    ATSS~\cite{ATSS}                    & 82.8            & 98.0         & 91.4         \\
    FCOSv2~\cite{fcosv2}                  & 80.0            & 98.6         & 89.1        \\
    Faster R-CNN~\cite{faster-rcnn}     & 73.7            & 92.5         & 83.3 \\
    \bottomrule
\end{tabular}

  \end{center}
  \vspace{-3mm}
  \caption{{\bf Detection comparison on YCB.} 
  Our method consistently outperforms other methods, especially in terms of AP$_{75}$.
  }
  \label{tab:ycbv_res50_compare}
\end{table}

\begin{figure}
    \centering
    \setlength\tabcolsep{1pt}
    \begin{tabular}{cc}
        \includegraphics[width=0.49\linewidth]{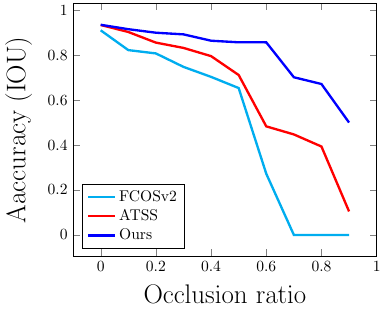} & 
        \includegraphics[width=0.49\linewidth]{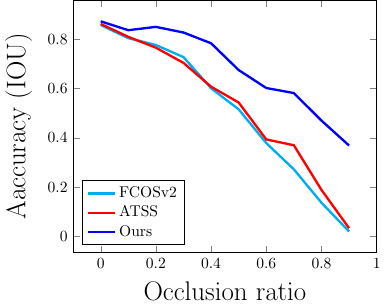} \\
        {\small (a) YCB}       & {\small (b) LM-O}
    \end{tabular}
    \caption{{\bf Performance w.r.t. different occlusion ratios.}
    Our method is much more robust to occlusions than the baselines.
    }
    \label{fig:occlusion_effect}
\end{figure}

\begin{table}[]
    \centering
    \begin{tabular}{lccc}
    \toprule
    Method                               & AP            & AP$_{50}$     & AP$_{75}$\\
    \midrule
    Center$^\dagger$      & 80.0          & 98.6 & 89.1 \\  
    Center                &  80.2         & 98.6  & 89.5\\
    {\bf Ours}$^\dagger$                   & 84.2              & 99.3          & 96.2\\
    {\bf Ours}               & \underline{85.0}  & \underline{99.4}& \underline{97.4}        \\
    {\bf Oracle$^\dagger$}             &84.8      & \textbf{99.6} & 97.2\\
    {\bf Oracle}              &\textbf{85.7}      & \textbf{99.6} &\textbf{97.9}\\
    \bottomrule
\end{tabular}
    \caption{{\bf Ablation study of different strategies on YCB.}
    We compare the center-based and the proposed visibility-guided sampling strategies used with standard NMS (denoted by $\dagger$) or with our fusion strategy. Our method is more accurate than the baseline strategies and performs on par with the oracle one that relies on guided sampling from the ground-truth mask. Here, ``Center$^\dagger$'' corresponds to the strategy of FCOSv2~\cite{fcosv2}.
    }
    \label{tab:ablation_study}
\end{table}

\begin{table}
    \centering
    \scalebox{0.95}{
    \begin{tabular}{cccc|cccc}
        \toprule
         $\mathcal{T}$     & AP        & AP$_{50}$     & AP$_{75}$  & $\alpha$ & AP        & AP$_{50}$     & AP$_{75}$ \\
        \midrule
         0.1       & 84.1      & 98.5          & 96.0           & 0        & 83.9          & 98.8          & 94.3 \\
       0.2       & 84.7      & {99.3}& 96.8          & 0.1       & {\bf 85.0}    & {\bf 99.4}    & {\bf 97.4}\\
        0.25    & {\bf 85.0}& {\bf 99.4}    & {\bf 97.4}    & 0.2       & {84.2} & {99.2}&{95.9}\\
        0.3       & {84.9}& {\bf 99.4}&{97.0}& 0.3       & 83.1          & 98.8          & 93.2\\
        0.4       & 84.0      & 98.6          & 96.6          & 0.4       & 81.2          & 98.0          & 90.8 \\
        \bottomrule
    \end{tabular}
    }
    \caption{{\bf Ablation study of different hyper-parameters on YCB.}}
    \label{tab:ablation_study_params}
\end{table}



\begin{table*}
    \centering

\begin{tabular}{lll|lllllll|l}
    \toprule
    Method                      & Real              & Data      & LM-O$^{\ast}$              & T-LESS             & TUD-L             & IC-BIN$^{\ast}$            & ITODD$^{\ast}$             & HB$^{\ast}$                & YCB             & Avg.          \\
    \midrule
    \textbf{PFA+Ours}           &                   & RGB       & \textbf{0.715}     & 0.719             & \textbf{0.733}    & \textbf{0.600}& 0.353              & \textbf{0.840}     & \textbf{0.648}    & \textbf{0.658} \\
    PFA~\cite{pfa}              &                   & RGB       & 0.674              &  -                & -                 &  -             & -                   & -              & 0.614             & -   \\
    SurfEmb~\cite{surfemb}      &                   & RGB       & 0.663              & \textbf{0.735}    & 0.715             &  0.588        & \textbf{0.413}     & 0.791               & 0.647             & 0.650 \\
    CosyPose~\cite{cosypose}    &                   & RGB       & 0.633              & 0.640             & 0.685             &  0.473        & 0.216              & 0.656              & 0.574             & 0.570 \\
    CDPNv2~\cite{cdpn}          &                   & RGB       & 0.624              & 0.407             & 0.588             &  0.226        & 0.067              & 0.722               & 0.390             & 0.472 \\
    \midrule
    \textbf{PFA+Ours}           & \checkmark        & RGB       & \textbf{0.715}& \textbf{0.778}& \textbf{0.839}   & \textbf{0.600}& 0.353          & \textbf{0.840}    & 0.806             & \textbf{0.704} \\
    PFA~\cite{pfa}              & \checkmark        & RGB       & 0.674    & -                 & -                 & -                     & -               & -               & 0.748             & -     \\
    SurfEmb~\cite{surfemb}      & \checkmark        & RGB       & 0.663    & 0.770             & 0.805             & 0.588        & \textbf{0.413} & 0.791    & 0.711             & 0.677             \\
    CosyPose~\cite{cosypose}    & \checkmark        & RGB       & 0.633    & 0.728             & 0.823             & 0.583        & 0.216    & 0.656    & \textbf{0.821}    & 0.637 \\
    CDPNv2~\cite{cdpn}          & \checkmark        & RGB       & 0.624    & 0.478             & 0.772             & 0.473        & 0.067     & 0.722     & 0.532             & 0.529 \\
    \midrule
    \textbf{PFA+Ours}           &                   & RGBD     & \textbf{0.797}    & 0.801             & \textbf{0.894}    & \textbf{0.676}    & 0.460             & \textbf{0.869}    & \textbf{0.826}    & \textbf{0.762} \\
    PFA~\cite{pfa}              &                   & RGBD     & 0.751             & -                 &  -                & -                 & -                 & -                 & 0.804             & -\\
    SurfEmb~\cite{surfemb}      &                   & RGBD     & 0.760             & \textbf{0.828}    & 0.854             & 0.659             & \textbf{0.538}    & 0.866             & 0.799             & 0.758 \\
    CDPNv2+ICP~\cite{cdpn}      &                   & RGBD     & 0.630             & 0.435             & 0.791             & 0.450             & 0.186             &  0.712            & 0.532             & 0.534 \\
    \midrule
    \textbf{PFA+Ours}           & \checkmark         & RGBD     & \textbf{0.797}    & \textbf{0.850}    & 0.960            & \textbf{0.676}    & 0.460    & \textbf{0.869}& 0.888             & \textbf{0.787} \\
    PFA~\cite{pfa}              & \checkmark        & RGBD     & 0.751    & -                 &  -                & -                 & -                 & -                 & 0.823             & -\\
    SurfEmb~\cite{surfemb}      & \checkmark        & RGBD     & 0.760    & 0.833             & 0.933             & 0.659    & \textbf{0.538}& 0.866& 0.824             & 0.773 \\
    CIR~\cite{coupled_iterative}& \checkmark        & RGBD     & 0.734             & 0.776             & \textbf{0.968}    & \textbf{0.676}    & 0.381             & 0.757             & \textbf{0.893}    & 0.741 \\
    CosyPose+ICP~\cite{cosypose}& \checkmark        & RGBD     & 0.714             & 0.701             & 0.939             & 0.647             & 0.313             & 0.712             & 0.861             & 0.698 \\
    CDPNv2+ICP~\cite{cdpn}      & \checkmark        & RGBD     & 0.630    & 0.464             & 0.913             & 0.450    & 0.186    & 0.712    & 0.619            & 0.568 \\
    \bottomrule
\end{tabular}

    \caption{\textbf{Comparison against the state of the art on 6D pose estimation.}
    Our detection method improves the original PFA-Pose by a large margin, and yields state-of-the-art results with either only synthetic or mixed data in both the RGB and RGBD settings. Note that LM-O, IC-BIN, ITODD, and HB provide only the synthetic PBR images for training, so the numbers ``w/o Real'' and ``w/ Real'' are the same on those datasets, indicated by ``$\ast$''. Here we report the results as the average of MSPD, MSSD, and VSD.
    }
    \label{tab:overall_compare}
\end{table*}

\begin{table}
  \centering
  \begin{tabular}{lcccc}
    \toprule
      Method          & Avg.      &  MSPD     & MSSD   & VSD  \\
    \midrule
    WDR + {\bf Ours}    & {\bf 0.605}  & {\bf 0.694}    & {\bf 0.598}    & {\bf 0.522}    \\
    WDR + RCNN          & 0.587             & 0.673             & 0.580         & 0.508\\
    WDR + FCOSv2          & 0.585             & 0.671            & 0.578        & 0.506\\
    \cmidrule{2-5}
    CDPNv2 + {\bf Ours} & {\bf 0.412}       & {\bf 0.534} & {\bf 0.428}   & {\bf 0.275}\\
    CDPNv2 + FCOSv2       & 0.402             & 0.523             & 0.416         & 0.268 \\
    CDPNv2 + RCNN       & 0.388             & 0.506             & 0.401         & 0.258\\
    \bottomrule
\end{tabular}
  \caption{
  \textbf{Effect on different pose regression networks.}
    Our detection method consistently improves the results of different pose regression frameworks, including WDR~\cite{wdr} and CDPNv2~\cite{cdpn}. Here we denote Mask R-CNN~\cite{maskrcnn} as ``RCNN'.
   }
  \label{tab:detect_for_two_stage}
  \vspace{-10pt}
\end{table}

\begin{figure*}[t]
  \centering
  \setlength\tabcolsep{1pt}
  
\begin{tabular}{cccc}
    \includegraphics[width=0.23\linewidth]{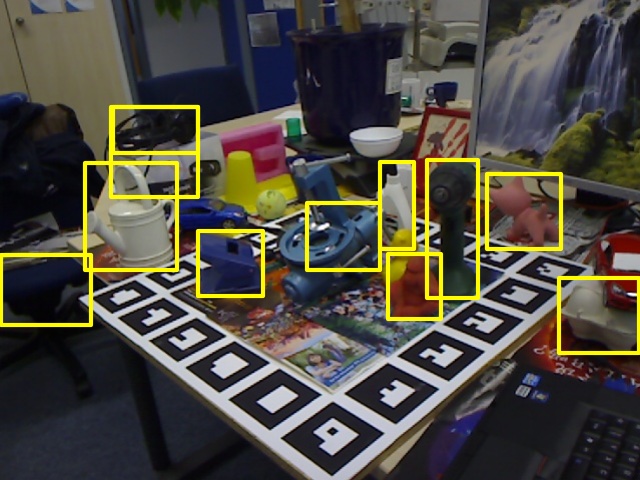} &
    \includegraphics[width=0.23\linewidth]{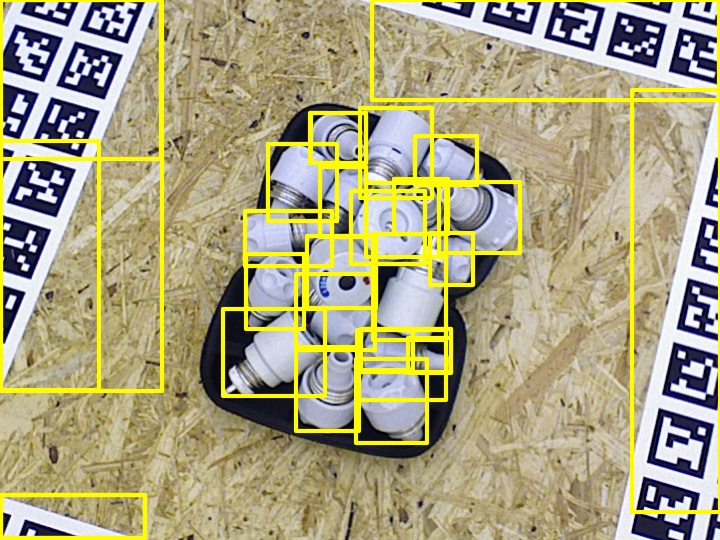} &
    \includegraphics[width=0.23\linewidth]{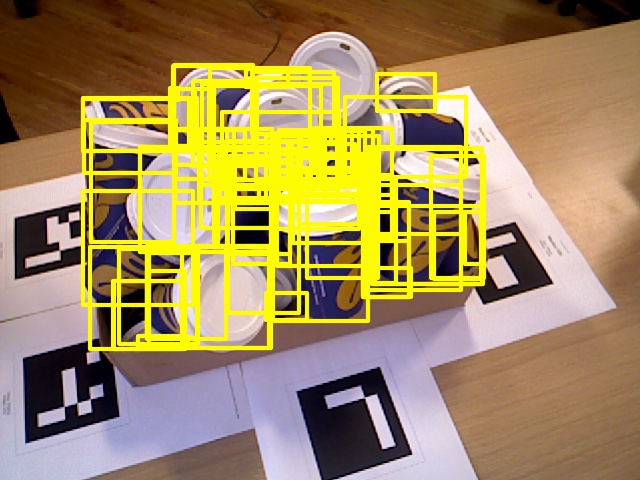} &
    \includegraphics[width=0.23\linewidth]{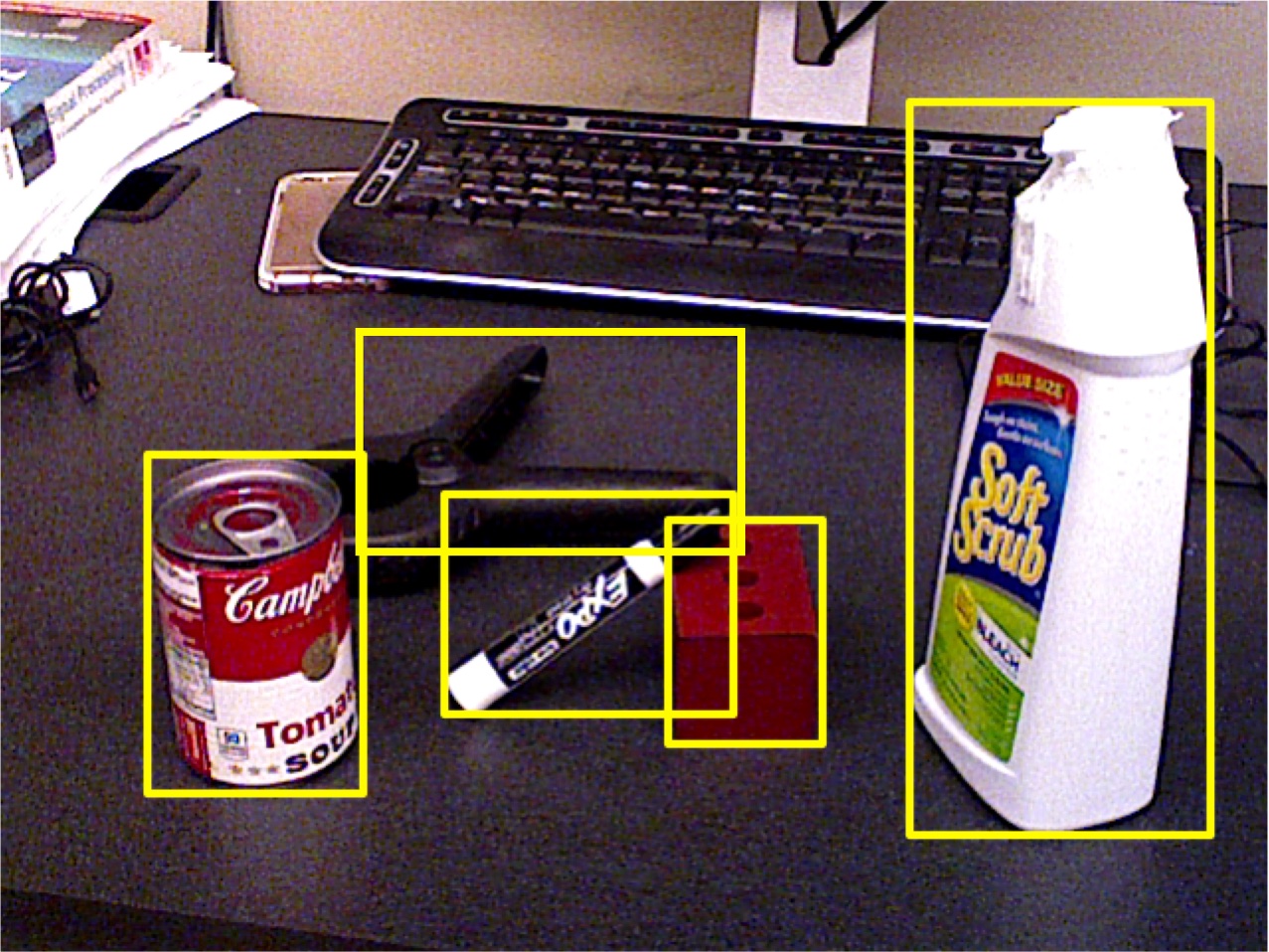} \\
    \includegraphics[width=0.23\linewidth]{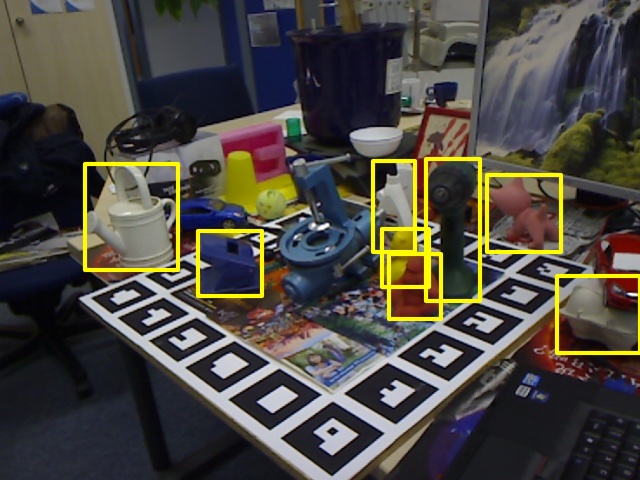} &
    \includegraphics[width=0.23\linewidth]{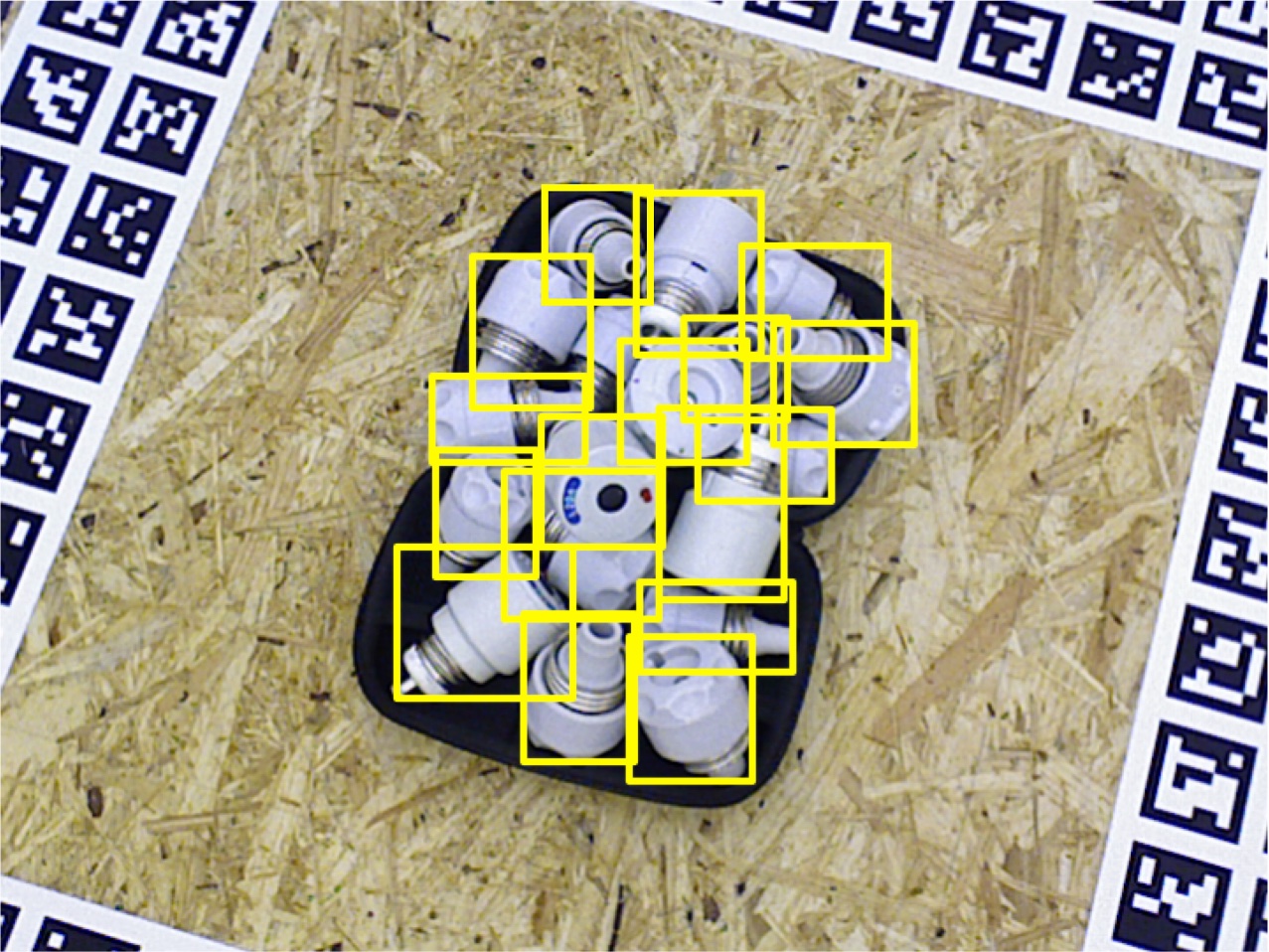} &
    \includegraphics[width=0.23\linewidth]{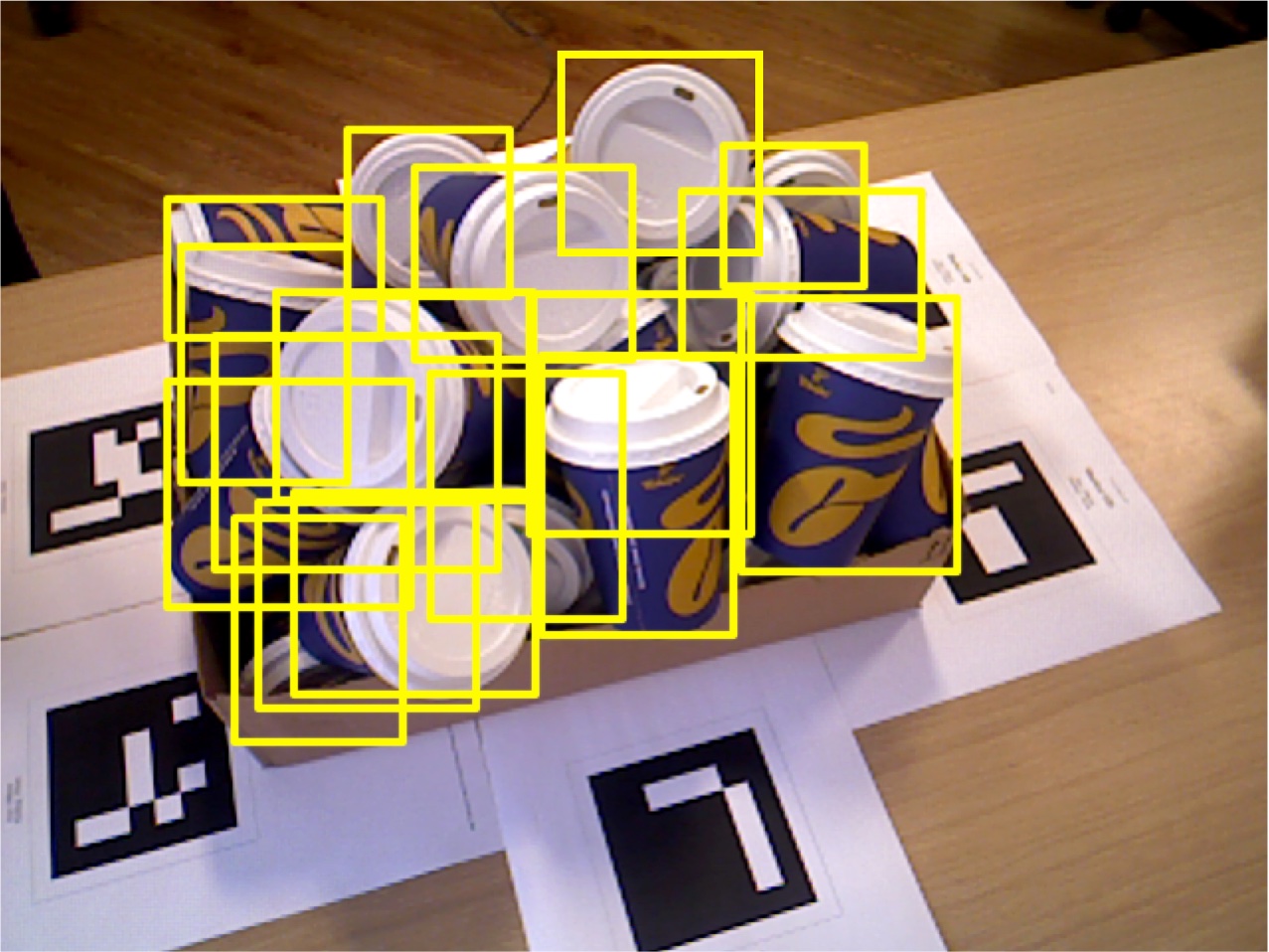} &
    \includegraphics[width=0.23\linewidth]{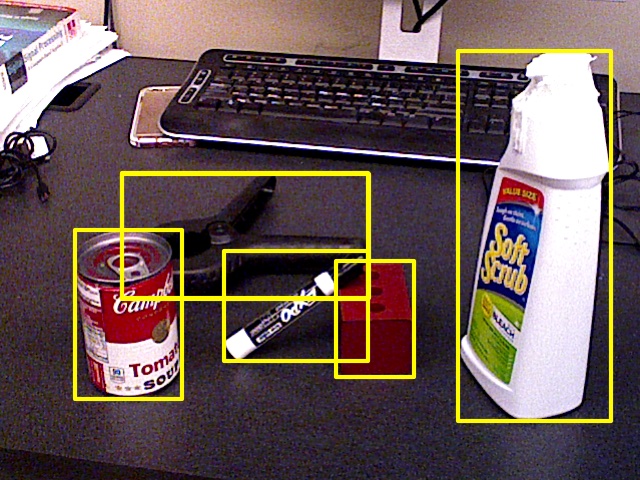} \\
    \includegraphics[width=0.23\linewidth]{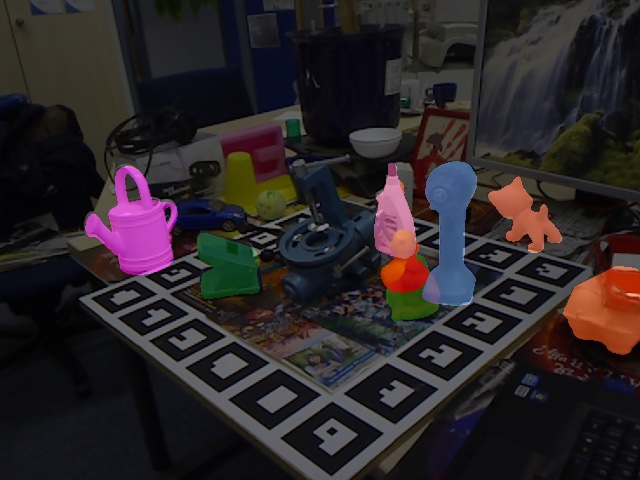} &
    \includegraphics[width=0.23\linewidth]{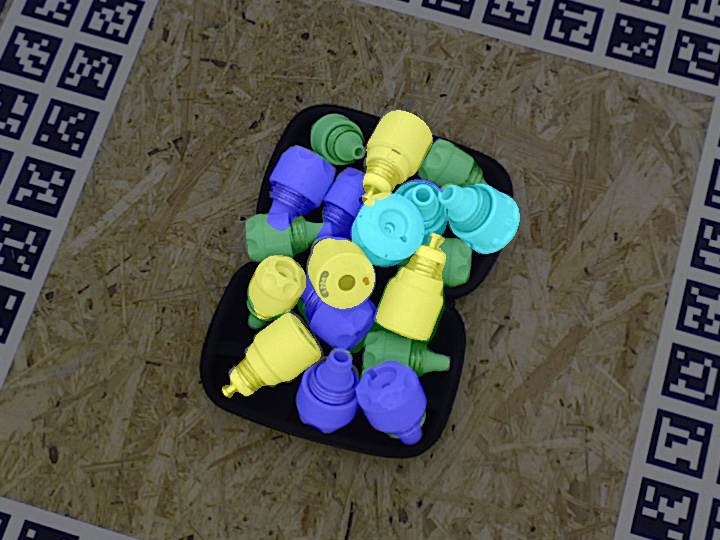} &
    \includegraphics[width=0.23\linewidth]{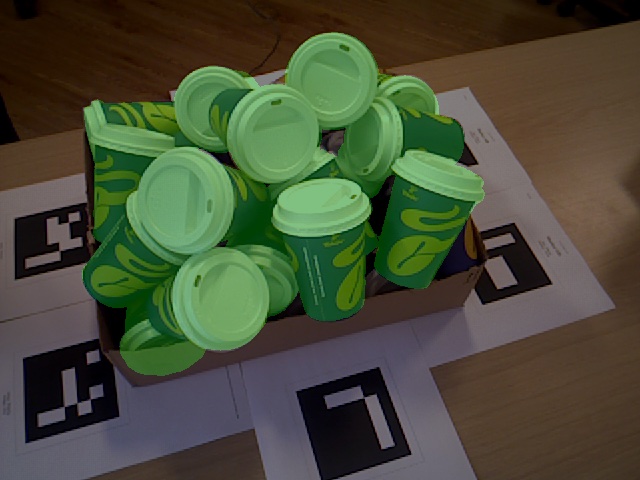} &
    \includegraphics[width=0.23\linewidth]{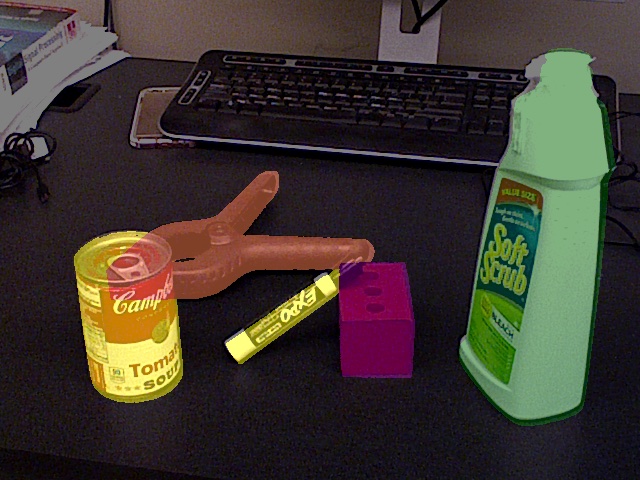} \\

\end{tabular}

  \caption{\textbf{Visualization of detection and pose results.}
  The first and second rows show the detection results of the baseline FCOS~\cite{fcosv1} and our method on different datasets (LM-O, T-LESS, IC-BIN, and YCB), respectively. Although the baseline works almost equally well in simple cases, such as targets without occlusions, it deteriorates significantly for targets in cluttered scenes, and generate many more false positives. By contrast, our detection method is robust, and produces accurate pose estimates after using a subsequent pose regression network (PFA~\cite{pfa}), as shown in the last row.
  }
  \label{fig:qualitative}
\end{figure*} 

\subsection{Object Detection}
\label{sec:exp_detection}

\noindent \textbf{Comparison with the baselines.}
We compare our method with the baseline single-stage method FCOSv2~\cite{fcosv2} and a typical two-stage method, Mask R-CNN~\cite{maskrcnn}.
As shown in Table~\ref{tab:detect_compare1}, our method outperforms them by a large margin on all datasets from the BOP benchmarks, demonstrating the effectiveness of our approach at detecting rigid objects in cluttered 6D pose estimation scenarios.

\noindent \textbf{Comparison with the state of the art on YCB.}
We compare our method with the state-of-the-art detection methods, including AutoAssign~\cite{autoassign} and PAA ~\cite{PAA}, on the YCB dataset. Table~\ref{tab:ycbv_res50_compare} summarizes the results, showing that our method consistently outperforms the state-of-the-art ones, especially in terms of AP$_{75}$.

\noindent \textbf{Performance under occlusions.}
Occlusion is a common problem in most BOP benchmarks.
We study the impact of different occlusion levels on different detectors. We compare our method with FCOSv2~\cite{fcosv2} and ATSS~\cite{ATSS} on both YCB and LM-O, and compute the average accuracy of the results with respect to the targets' occlusion ratio. The results are summarized in Fig.~\ref{fig:occlusion_effect}. Although ATSS improves the center-based sampling of FCOS by its adaptive assignment strategy across multiple pyramid levels, it remains sensitive to occlusions, as illustrated by the quick deterioration of accuracy with the increasing of occlusion ratio. By contrast, our method is much more robust.

\noindent \textbf{Ablation study on YCB.} 
We compare NMS and our fusion strategy on a model trained either with the centered-based strategy or the proposed visibility-guided one.
As shown in Table~\ref{tab:ablation_study}, with the same NMS post-processing, our sampling strategy already outperforms the center-based baseline by 4.2 points. This confirms the importance of involving all the visible object parts during training, leveraging the rigidity of the targets. Furthermore, both sampling strategies benefit from our fusion method discussed in Section~\ref{sec:fusion}. However, it only increases the performance of the center-based one by 0.2 points, which highlights the drawback of not using non-center areas during training, making the fusion during inference less effective. By contrast, our fusion method increases the performance of our sampling strategy by 0.8 points, making it perform on par with the oracle that uses the ground-truth mask to guide sampling.

Furthermore, we evaluate the performance of our method with different $\mathcal{T}$ and $\alpha$, where $\mathcal{T}$ is the threshold to filter out easy negative cells before sampling positive ones, and $\alpha$ is the weight balancing the Euclidean distance and the minimum barrier distance in Eq.~\ref{eqn:mbd}.
We use the default settings of $\mathcal{T}=0.25$ and $\alpha=0.1$, and vary only one parameter at a time. As shown in Table~\ref{tab:ablation_study_params}, the results are stable when $\mathcal{T}$ is set between 0.2 and 0.3, and we use the best value 0.25 found on YCB for all our experiments. Following~\cite{minbarrier}, we use the default value 0.1 for $\alpha$, which is further supported by our results with different $\alpha$ values.

\noindent \textbf{Runtime analysis.}
We conduct all our experiments on a workstation with an NVIDIA RTX-3090 GPU and an Intel-Xeon CPU with 12 2.1GHz cores. Our method shares the same network architecture as most single-stage methods~\cite{fcosv2,ATSS,PAA, autoassign} and the running time of our simple fusion strategy is negligible. As such, all methods have a similar inference speed of about 32.4 images per second on the YCB dataset with an average of 4.8 instances in each image. The main difference comes from the training time, since different methods rely on different sampling strategies. Our method has a throughput of about 18.7 images per second during training, which is slightly slower than FCOSv2 (22.3) and ATSS (21.6), but faster than PAA (16.6) and AutoAssign (15.7).

\subsection{Object Pose Estimation}
\label{sec:exp_pose}

\noindent \textbf{Comparison with the state of the art.}
To demonstrate the effectiveness of our detection method in 6D object pose estimation, we combine it with a recent pose regression network, PFA-Pose~\cite{pfa}, and compare the pose results with other methods. We test our method with PFA-Pose in different settings, including training only on synthetic PBR or with mixed real images. Additionally, we evaluate our method when PFA-Pose uses a simple depth refinement strategy based on RANSAC-Kabsch~\cite{dpodv2,ncf} to consume additional depth images. The original PFA-Pose cannot handle multiple instances from the same class, making it inapplicable to some datasets. So we only reproduce its results on LM-O and YCB. Table~\ref{tab:overall_compare} summarizes the results, showing that our detection method improves the original PFA-Pose by a large margin, obtaining state-of-the-art pose estimation results with either only synthetic or mixed data in both the RGB and RGBD settings. Fig.~\ref{fig:qualitative} visualizes some results.

\noindent \textbf{Evaluation with different pose regression networks.}
In principle, our detection method can be used with most pose regression frameworks as a first component to extract the object's bounding box before pose regression.
To demonstrate its generalization ability, we test our detection method on YCB with two other typical pose regression networks, WDR-Pose~\cite{wdr} and CDPNv2~\cite{cdpn}. 
Table~\ref{tab:detect_for_two_stage} provides the results, evidencing that our detection method consistently improves the pose results.

\section{Conclusion}
We have proposed a visibility-guided sampling strategy for training a deep network to detect rigid objects in cluttered scenes.
We first analyzed the influence of the rigidity of the targets in the 6D object pose estimation scenarios and studied the weaknesses of general detection methods in this setting. Based on the observation that detecting rigid objects should allow us to rely on all visible object parts and that each part should already provide a reliable prediction of the whole bounding box, we have proposed to build a visibility map to guide the positive sampling during training and combine multiple local predictions during inference to obtain the final robust result. We have demonstrated the effectiveness of our method on the challenging datasets from the BOP benchmarks. It achieves much better detection results than general methods and produces state-of-the-art pose results when combined with pose regression networks.


In the future, we will seek to use learning-based strategies to model the visibility of object parts without mask annotations, and investigate better fusion strategies to obtain robust detection results from local predictions.

\vspace{0.2em}
{\small
{\noindent \bf Acknowledgments.}
This work was supported by the 111 Project of China under Grant B08038, the Fundamental Research Funds for the Central Universities under Grant JBF220101, and the Youth Innovation Team of Shaanxi Universities. We thank Zhaoyang Liu and Wayne Wu for helpful discussions.
}


\begin{table*}
    \centering
    \begin{tabular}{l@{\hspace{3em}}ccc@{\hspace{3em}}ccc@{\hspace{3em}}ccc}
        \toprule
        \multirow{2}{*}{Category}  & \multicolumn{3}{c@{\hspace{3em}}}{FCOSv2~\cite{fcosv2}}                &\multicolumn{3}{c@{\hspace{3em}}}{PAA~\cite{PAA}}             &\multicolumn{3}{c}{\bf Ours}         \\
        ~                           & AP        & AP$_{50}$     & AP$_{75}$     & AP            & AP$_{50}$     & AP$_{75}$ & AP            & AP$_{50}$     & AP$_{75}$ \\
        \midrule
        Toaster                     & {20.9} &{38.4}& {30.5}          & 16.1          & 34.0          & 25.7      & {\bf 31.8}    & {\bf 47.2}    & {\bf 42.3}         \\
        Bottle                      & {38.3}      & {59.4}  & {43.1}          & 37.0          & 58.5          & 41.9        & {\bf 40.5}  & {\bf 60.7}    & {\bf 45.7}     \\
        Skateboard                  & 50.1       & {69.7} & 59.4          & {50.2}  & 69.3  & {60.2}      & {\bf 52.6} & {\bf 71.5}  & {\bf 63.8}        \\
        Suitcase                    & 34.7      & {52.5}          & 38.8          & {35.1} & 52.0   & {39.0}       & {\bf 36.5}   & {\bf 53.6}    & {\bf 40.7}        \\ 
        Cup                         & 41.3      & 62.3  & {50.6}  & {41.5}  & {62.7}  & {50.6}      & {\bf 42.8}    & {\bf 63.2}    & {\bf 52.2}        \\
        ... & & ... & & & ... & & & ... \\
        \midrule
        Cat                         & {65.7}  & {86.8} & {68.4}          & {\bf 67.2}   & {\bf 88.9} & {\bf 69.8}     & 63.0          & 84.5          & 66.2        \\
        Dog                         & {60.7} & {81.9}   & {62.4}          & {\bf 62.1}    & {\bf 83.2}    & {\bf 63.9}       & 59.3          & 80.8         & 60.7     \\  
        Cow                         & {56.4}      & {72.3}  & {60.5}          & {\bf 58.0}    & {\bf 74.0}    & {\bf 62.8}      & 56.0           & 71.0       & 60.2   \\
        Sheep                       & {51.4} & {71.2}  & {60.7}          & {\bf 51.7}   & {\bf 71.6} & {\bf 61.2}      & 50.0          & 70.3          & 58.6         \\
        Bird                        & {\bf 36.2}& {\bf 54.8}& {\bf 42.3}    & {35.6}  & {54.0}  & 41.4     & 35.1          & 53.5          & {41.6}      \\
        ... & & ... & & & ... & & & ... \\
        \midrule
        {Avg.}                  & 38.9      & \underline{57.5}  & 42.2          & {\bf 40.4}          & {\bf 58.4}          & {\bf 43.9}      & \underline{40.0}    & 57.1  & \underline{43.4}      \\
        \bottomrule
    \end{tabular}
    \caption{{\bf Evaluation of general detection on COCO.} Although the general detection dataset COCO does not fully match our assumption of rigid targets, our method outperforms the baselines significantly on categories such as toaster, bottle, etc., which are mainly rigid objects, and our method achieves similar performance to the baselines in average accuracy.
    }
    \label{tab:coco_eval}
\end{table*}

\section{Appendix}

\noindent \textbf{General scenario.} Our work is motivated by the rigidity of the targets in 6D object pose estimation. The general scenario, e.g., COCO, does not fully match our assumption. Nevertheless, we report results with the same experimental setting as FCOSv2 and PAA in Table~\ref{tab:coco_eval}. In addition to the average accuracy across the 80 COCO categories, we report the accuracy of the 5 categories on which our method outperforms the baselines the most, and the 5 categories on which our method performs the worst. Our method outperforms the baselines significantly on categories such as toaster, bottle, etc., which are mainly rigid objects. By contrast, the categories on which our method underperforms include cat, dog, etc., which are mainly non-rigid targets and break our assumption. Our method nevertheless achieves similar performance to the baselines in average accuracy.


\noindent \textbf{Additional quantitative results.}
We show the detailed object pose results using different metrics on LM-O, T-LESS, TUD-L, IC-BIN, ITODD, HB, and YCB in Table~\ref{tab:compare_lmo},~\ref{tab:compare_tless},~\ref{tab:compare_tudl},~\ref{tab:compare_icbin},~\ref{tab:compare_itodd},~\ref{tab:compare_hb}, and~\ref{tab:compare_ycbv}, respectively.
Our method combined with PFA-Pose~\cite{pfa} outperforms the state of the art in most experimental settings.

\begin{table}[h]
    \centering
    \scalebox{0.95}{
    \begin{tabular}{lcccc}
    \toprule
    Method  &  Avg.     & MSPD      & MSSD      & VSD\\
    \midrule
    \multicolumn{5}{c}{\textit{RGB (PBR)}} \\
    \midrule
    {\bf PFA+Ours}  &{\bf 0.715}&{\bf 0.876}&{\bf 0.712}&{\bf 0.559} \\
    PFA             & 0.674     & 0.819     & 0.673     & 0.531 \\
    SurfEmb         & 0.663     & 0.851     & 0.649     & 0.497 \\
    CIR             & 0.655     & 0.831     & 0.633     & 0.501 \\
    Cosypose        & 0.633      & 0.812     & 0.606     & 0.480 \\
    CDPNv2          & 0.624     & 0.815     & 0.612     & 0.445 \\
    \midrule
    \multicolumn{5}{c}{\textit{RGB-D (PBR)}} \\
    \midrule
    {\bf PFA+Ours}  &{\bf 0.797}&{\bf 0.890}&{\bf 0.712}&{\bf 0.559} \\
    PFA             & 0.751     & 0.835     & 0.673     & 0.531 \\
    SurfEmb         & 0.760     & 0.856     & 0.649     & 0.497 \\
    CIR             & 0.734     & 0.824     & 0.633     & 0.501 \\
    Cosypose+ICP    & 0.714     & 0.826     & 0.606     & 0.480 \\
    CDPNv2+ICP      & 0.630     & 0.731     & 0.612     & 0.445 \\
    \bottomrule
\end{tabular}
    }
    \caption{{\bf Additional object pose results on LM-O.}
    }
    \label{tab:compare_lmo}
\end{table}

\begin{table}
    \centering
     \scalebox{0.95}{
    \begin{tabular}{lcccc}
    \toprule
    Method          & Avg.     & MSPD        & MSSD      & VSD\\
    \midrule
    \multicolumn{5}{c}{\textit{RGB (PBR)}}\\
    \midrule
    {\bf PFA+Ours}  & 0.719     & 0.832     & 0.682     & 0.643 \\
    SurfEmb         &{\bf 0.735}&{\bf 0.857}&{\bf 0.686}& {\bf 0.661} \\
    Cosypose        & 0.640      & 0.761     & 0.589     & 0.571 \\
    CDPNv2          & 0.407      & 0.579     & 0.338     & 0.303 \\
    \midrule
    \multicolumn{5}{c}{\textit{RGB (Real+PBR)}}\\
    \midrule
    {\bf PFA+Ours}  &{\bf 0.778}&{\bf 0.877}&{\bf 0.749}&{\bf 0.709} \\
    SurfEmb         &  0.770    & -         & -         & - \\
    CIR             &  0.715    & 0.798     & 0.684     & 0.663\\
    Cosypose        &  0.728    & 0.821     & 0.695     & 0.669 \\
    CDPNv2          &  0.478    & 0.620     & 0.426     & 0.386 \\
    \midrule
    \multicolumn{5}{c}{\textit{RGB-D (PBR)}}\\
    \midrule
    {\bf PFA+Ours}  & 0.801     & 0.833     & 0.807     & 0.764 \\
    SurfEmb         &{\bf 0.828}&{\bf 0.859}&{\bf 0.829}& {\bf 0.797} \\
    CDPNv2+ICP      & 0.435     & 0.488     & 0.449     & 0.368 \\
    \midrule
    \multicolumn{5}{c}{\textit{RGB-D (Real+PBR)}}\\
    \midrule
    {\bf PFA+Ours}  &{\bf 0.850}&{\bf 0.878}&{\bf 0.856}&{\bf 0.816} \\
    SurfEmb         & 0.833     & -         & -         & - \\
    CIR             & 0.776     & 0.795     & 0.773     & 0.760\\
    Cosypose+ICP    & 0.701     & 0.767     & 0.749     & 0.587 \\
    CDPNv2+ICP      & 0.464     & 0.516     & 0.489     & 0.385 \\
    \bottomrule
\end{tabular}
    }
    \caption{{\bf Additional object pose results on T-LESS.}
    }
    \label{tab:compare_tless}
\end{table}

\begin{table}
    \centering
     \scalebox{0.95}{
    \begin{tabular}{lcccc}
    \toprule
    Method          & Avg.     & MSPD        & MSSD      & VSD\\
    \midrule
    \multicolumn{5}{c}{\textit{RGB (PBR)}}\\
    \midrule
    {\bf PFA+Ours}  &{\bf 0.733}&{\bf 0.890}&{\bf 0.721}& {\bf 0.594} \\
    SurfEmb         & 0.715     & 0.889     & 0.687     & 0.569 \\
    Cosypose        & 0.685     & 0.847     & 0.664    & 0.544 \\
    CDPNv2          & 0.588     & 0.797     & 0.577     & 0.391 \\
    \midrule
    \multicolumn{5}{c}{\textit{RGB (Real+PBR)}}\\
    \midrule
    {\bf PFA+Ours}  &{\bf 0.839}&{\bf 0.978}&{\bf 0.820}&{\bf 0.719} \\
    SurfEmb         & 0.805     & -         & -         & - \\
    Cosypose        & 0.823     & 0.973     &0.807      &0.689 \\
    CDPNv2          & 0.772     & 0.925     & 0.793     & 0.597 \\
    \midrule
    \multicolumn{5}{c}{\textit{RGB-D (PBR)}}\\
    \midrule
    {\bf PFA+Ours}  &{\bf 0.894}&{\bf 0.929}&{\bf 0.930}& {\bf 0.821} \\
    SurfEmb         & 0.854     & 0.905     & 0.891     & 0.767 \\
    CDPNv2+ICP      & 0.791     & 0.829     & 0.847     & 0.698 \\
    \midrule
    \multicolumn{5}{c}{\textit{RGB-D (Real+PBR)}}\\
    \midrule
    {\bf PFA+Ours}  & 0.960     & 0.989     & 0.986     & 0.904 \\
    CIR             &{\bf 0.968}&{\bf 0.991}&{\bf 0.991}&{\bf 0.920}\\
    SurfEmb         & 0.933     & -         & -         & - \\
    Cosypose+ICP    & 0.939     & 0.976     & 0.972     & 0.869 \\
    CDPNv2+ICP      & 0.913     & 0.947     & 0.962     & 0.832 \\
    \bottomrule
\end{tabular}
    }
    \caption{{\bf Additional object pose results on TUD-L.}
    }
    \label{tab:compare_tudl}
\end{table}

\begin{table}
    \centering
     \scalebox{0.95}{
    \begin{tabular}{lcccc}
    \toprule
    Method  &  Avg.     & MSPD      & MSSD      & VSD\\
    \midrule
    \multicolumn{5}{c}{\textit{RGB (PBR)}} \\
    \midrule
    {\bf PFA+Ours}  &{\bf 0.600}&{\bf 0.689}& 0.589     &{\bf 0.521}\\
    SurfEmb         & 0.588     & 0.678     &{\bf 0.573}& 0.514 \\
    Cosypose        & 0.473     & 0.675     & 0.559     & 0.515 \\
    CDPNv2          & 0.226     & 0.582     & 0.438     & 0.399 \\
    \midrule
    \multicolumn{5}{c}{\textit{RGB-D (PBR)}} \\
    \midrule
    {\bf PFA+Ours}  &{\bf 0.676}&{\bf 0.702}&{\bf 0.692}& 0.636 \\
    SurfEmb         & 0.659     & 0.680     &0.677      & 0,621 \\
    CIR             &{\bf 0.676}& 0.683     & 0.688     &{\bf 0.656} \\
    Cosypose+ICP    & 0.647     & 0.666     & 0.652     & 0.624 \\
    CDPNv2+ICP      & 0.450     & 0.459     & 0.458     & 0.433 \\
    \bottomrule
\end{tabular}
    }
    \caption{{\bf Additional object pose results on IC-BIN.}
    }
    \label{tab:compare_icbin}
\end{table}

\begin{table}
    \centering
     \scalebox{0.95}{
    \begin{tabular}{lcccc}
    \toprule
    Method  &  Avg.     & MSPD      & MSSD      & VSD\\
    \midrule
    \multicolumn{5}{c}{\textit{RGB (PBR)}} \\
    \midrule
    {\bf PFA+Ours}  & 0.353     & 0.484     & 0.306     & 0.269 \\
    SurfEmb         &{\bf 0.413}&{\bf 0.552}&{\bf 0.363}&{\bf 0.324} \\
    Cosypose        & 0.216     & 0.300     & 0.177     & 0.172 \\
    CDPNv2          & 0.067     & 0.161     & 0.087     & 0.059 \\
    \midrule
    \multicolumn{5}{c}{\textit{RGB-D (PBR)}} \\
    \midrule
    {\bf PFA+Ours}  & 0.460     & 0.498     & 0.495     & 0.413 \\
    SurfEmb         &{\bf 0.538}&{\bf 0.560}&{\bf 0.558}&{\bf 0.497} \\
    CIR             & 0.381     & 0.370     & 0.379     & 0.394 \\
    Cosypose+ICP    & 0.313     & 0.315     & 0.341     & 0.282 \\
    CDPNv2+ICP      & 0.186     & 0.184     & 0.206     & 0.168 \\
    \bottomrule
\end{tabular}
    }
    \caption{{\bf Additional object pose results on ITODD.}
    }
    \label{tab:compare_itodd}
\end{table}

\begin{table}
    \centering
    \scalebox{0.95}{
    \begin{tabular}{lcccc}
    \toprule
    Method  &  Avg.     & MSPD      & MSSD      & VSD\\
    \midrule
    \multicolumn{5}{c}{\textit{RGB (PBR)}} \\
    \midrule
    {\bf PFA+Ours}  &{\bf 0.840}& 0.879     & 0.840     &{\bf 0.804}\\
    SurfEmb         & 0.791     &{\bf 0.888}&{\bf 0.760}& 0.725 \\
    Cosypose        & 0.656     & 0.721     & 0.634     & 0.613 \\
    CDPNv2          & 0.722     & 0.845     & 0.708     & 0.614 \\
    \midrule
    \multicolumn{5}{c}{\textit{RGB-D (PBR)}} \\
    \midrule
    {\bf PFA+Ours}  &{\bf 0.869}& 0.888     &{\bf 0.879}& {\bf 0.839} \\
    SurfEmb         & 0.866     &{\bf 0.893}&0.875      & 0.829 \\
    CIR             &{\bf 0.757}& 0.757     & 0.753     & 0.760 \\
    Cosypose+ICP    & 0.712     & 0.737     & 0.717     & 0.679 \\
    CDPNv2+ICP      & 0.712     & 0.749     & 0.757     & 0.629 \\
    \bottomrule
\end{tabular}
    }
    \caption{{\bf Additional object pose results on HB.}
    }
    \label{tab:compare_hb}
\end{table}

\begin{table}
    \centering
    \scalebox{0.95}{
    \begin{tabular}{lcccc}
    \toprule
    Method          & Avg.     & MSPD        & MSSD      & VSD\\
    \midrule
    \multicolumn{5}{c}{\textit{RGB (PBR)}}\\
    \midrule
    {\bf PFA+Ours}  &{\bf 0.648}& 0.771     &{\bf 0.623}& {\bf 0.550} \\
    PFA             & 0.614     & 0.739     & 0.585     & 0.522 \\
    SurfEmb         & 0.647      &{\bf 0.773}& 0.620     & 0.548 \\
    Cosypose        & 0.574     & 0.653     & 0.554     & 0.516 \\
    CDPNv2          & 0.390     & 0.511     & 0.399     & 0.260 \\
    \midrule
    \multicolumn{5}{c}{\textit{RGB (Real+PBR)}}\\
    \midrule
    {\bf PFA+Ours}  & 0.806     &{\bf 0.856}& 0.809     & 0.751 \\
    PFA             & 0.748     & 0.810     & 0.742     & 0.694 \\
    CIR             &{\bf 0.824}& 0.852     & 0.835     & {\bf 0.783}\\
    SurfEmb         & 0.711     & -         & -         & - \\
    Cosypose        & 0.821     & 0.850     &{\bf 0.842}& 0.772 \\
    CDPNv2          & 0.532     & 0.631     & 0.570     & 0.396 \\
    \midrule
    \multicolumn{5}{c}{\textit{RGB-D (PBR)}}\\
    \midrule
    {\bf PFA+Ours}  &{\bf 0.826}&{\bf 0.819}&{\bf 0.867}& {\bf 0.792} \\
    PFA             & 0.804     & 0.793     & 0.842     & 0.775 \\
    SurfEmb         & 0.799     & 0.792     & 0.849     & 0.757 \\
    CDPNv2+ICP      & 0.532     & 0.483     & 0.603     & 0.511 \\
    \midrule
    \multicolumn{5}{c}{\textit{RGB-D (Real+PBR)}}\\
    \midrule
    {\bf PFA+Ours}  & 0.888     & 0.881     & 0.920     & 0.863 \\
    PFA             & 0.823     & 0.816     & 0.852     & 0.803 \\
    CIR             &{\bf 0.893}&{\bf 0.885}& {\bf 0.924}& {\bf 0.871}\\
    SurfEmb         & 0.824       & -         & -         & - \\
    Cosypose+ICP    & 0.861     & 0.849     & 0.903     & 0.831 \\
    CDPNv2+ICP      & 0.619     & 0.565     & 0.701     & 0.590 \\
    \bottomrule
\end{tabular}
    }
    \caption{{\bf Additional object pose results on YCB.}
    }
    \label{tab:compare_ycbv}
\end{table}

{\small
\bibliographystyle{ieee_fullname}
\bibliography{egbib}
}

\end{document}